\documentclass{acmart}

\usepackage{hyperref}
\usepackage{graphicx}
\usepackage{amsmath}
\usepackage{array}
\usepackage{listings, xcolor}

\definecolor{verylightgray}{rgb}{.97,.97,.97}

\lstdefinelanguage{Solidity}{
	keywords=[1]{anonymous, assembly, assert, balance, break, call, callcode, case, catch, class, constant, continue, constructor, contract, debugger, default, delegatecall, delete, do, else, emit, event, experimental, export, external, false, finally, for, function, gas, if, implements, import, in, indexed, instanceof, interface, internal, is, length, library, log0, log1, log2, log3, log4, memory, modifier, new, payable, pragma, private, protected, public, pure, push, require, return, returns, revert, selfdestruct, send, solidity, storage, struct, suicide, super, switch, then, this, throw, transfer, true, try, typeof, using, value, view, while, with, addmod, ecrecover, keccak256, mulmod, ripemd160, sha256, sha3}, 
	keywordstyle=[1]\color{blue}\bfseries,
	keywords=[2]{address, bool, byte, bytes, bytes1, bytes2, bytes3, bytes4, bytes5, bytes6, bytes7, bytes8, bytes9, bytes10, bytes11, bytes12, bytes13, bytes14, bytes15, bytes16, bytes17, bytes18, bytes19, bytes20, bytes21, bytes22, bytes23, bytes24, bytes25, bytes26, bytes27, bytes28, bytes29, bytes30, bytes31, bytes32, enum, int, int8, int16, int24, int32, int40, int48, int56, int64, int72, int80, int88, int96, int104, int112, int120, int128, int136, int144, int152, int160, int168, int176, int184, int192, int200, int208, int216, int224, int232, int240, int248, int256, mapping, string, uint, uint8, uint16, uint24, uint32, uint40, uint48, uint56, uint64, uint72, uint80, uint88, uint96, uint104, uint112, uint120, uint128, uint136, uint144, uint152, uint160, uint168, uint176, uint184, uint192, uint200, uint208, uint216, uint224, uint232, uint240, uint248, uint256, var, void, ether, finney, szabo, wei, days, hours, minutes, seconds, weeks, years},	
	keywordstyle=[2]\color{teal}\bfseries,
	keywords=[3]{block, blockhash, coinbase, difficulty, gaslimit, number, timestamp, msg, data, gas, sender, sig, value, now, tx, gasprice, origin},	
	keywordstyle=[3]\color{violet}\bfseries,
	identifierstyle=\color{black},
	sensitive=false,
	comment=[l]{//},
	morecomment=[s]{/*}{*/},
	commentstyle=\color{gray}\ttfamily,
	stringstyle=\color{red}\ttfamily,
	morestring=[b]',
	morestring=[b]"
}

\AtBeginDocument{%
  \providecommand\BibTeX{{%
    \normalfont B\kern-0.5em{\scshape i\kern-0.25em b}\kern-0.8em\TeX}}}

\setcopyright{acmcopyright}
\copyrightyear{2020}
\acmYear{2020}
\acmDOI{10.1145/1122445.1122456}

\begin{document}

\title{Ethereum  Data  Structures}

\author{Kamil Jezek}
\email{kamil.jezek@sydney.edu.au}
\affiliation{%
  \institution{University of Sydney}
  \state{Australia}
}

\renewcommand{\shortauthors}{Kamil Jezek, et al.}

\begin{abstract}
Ethereum platform operates with rich spectrum of data structures and hashing and coding functions. The main source describing them is the Yellow paper, complemented by a lot of informal blogs. These sources are somehow limited. In particular, the Yellow paper does not ideally balance brevity and detail, in some parts it is very detail, while too shallow elsewhere. The blogs on the other hand are often too vague and in certain cases contain incorrect information. As a solution, we provide this document, which summarises data structures used in Ethereum. The goal is to provide sufficient detail while keeping brevity. Sufficiently detailed formal view is enriched with examples to extend on clarity.
\end{abstract}



\maketitle

\section{Introduction}

Blockchain servers as an automatic distributed ledger which comprise all transactions and state. While the blocks in the blockchain are executed, it continuously modifies the state, changes balances of accounts, produce transaction receipts etc. This system is non-stopping, accepting new transactions, creating new blocks and always growing the state. To handle this complexity and especially the growing volume of data, the blockchain architects re-used fundamental data structures known throughout the history of computer science, but combined them in new forms.

Since the data structures used in blockchain platforms are sometimes complex and sometimes specific only to  Ethereum, it is crucial to understand them to be able to comprehend the platform as such. This is especially important for researches and practitioners willing to extend, modify, build on top of, conduct a research on, etc. on Ethereum.  In this area two main sources of knowledge exist, first is the official documentation formulated in the Yellow paper \cite{wood2019ethereum} and the other one is endless chain of blogs on the Internet. 

We believe that available sources of information related especially to data-structures are limited. 
Reasonable body of publications cover blockchain and Ethereum as a platform from wider perspective. Let us name work done by Kolb at al \cite{kolb2020core}. His work introduces beginners into blockchain, covers aspects that we do not such as mining, incentivisation, etc, includes comparison of various blockchain platforms. However, when it comes into data structures, it is rather shallow, providing no better detail then online blogs.  Even wider perspective is provided by Lao at al \cite{lao2020survey} or by Zhu at al \cite{zhu2020applications}. These two papers survey suitability of blockchain as a tamper-proof storage for trust in internet of things. Both these papers to a limited degree overview structures of blockchains. More from the security point of view, the blockchain platforms are discussed by Zhang at al \cite{zhang2019security}.

Two student thesis, one by Voulgari \cite{Voulgari2019ethereum} and the other one by Hefele \cite{Hefele2019conceptual} are rich in describing Ethereum data structures. Especially Hefele's work is relevant as it model Ethereum data structures using UML diagrams.  This, according to our judgement, may bring clarity in complicated Ethereum data structures. Both works, however, describe data structures just as a means for analysing volume of data stored in these structures. The structures themselves are not described into the level that would show how the data are encoded in memory or disk. 

Rich information is provided by Antonopoulos and Wood, founders of Ethereum, in their book \cite{antonopoulos2018mastering}, or in another book by Bashir \cite{bashir2017mastering}. Both publications seem to target mainly end users of Ethereum as they detail all surrounding tools, but do not go deep into internals. 

For the all mentioned above, the only thorough source of information remains the Yellow paper\cite{wood2019ethereum}. We believe that this publication is not ideal for understanding Ethereum data structures. In particular, it scatters information across the publication with details about data encoding provided in appendixes, where it is sometimes not clear how partial parts are related. Moreover, this publication omit information of how Smart Contract data are structured and stored in memory slots. While they do it on purpose as the data format is left on the Smart Contract compiler, we believe that this part is crucial for understanding data organization of Ethereum. For this reason, we have also studied the Solidity compiler to provide the data format, and summarise it in this publication. 

To fill the gap of missing or unclear information from various parts of the Yellow paper, we had to rely on the internet blogs. Their problem is however that they  vary widely in quality as they rather capture people understanding. To cope with this limitation we have to in some cases consult implementation in source-code of Ethereum clients, namely Parity\cite{parity} and Geth\cite{geth}, to confirm certain details. We aim at providing sufficient detail without needlessly formal approach as it is the case in the Yellow paper, but we aim at using better detail than provided in a lot of blogs. To extend on clarity, we provide a lot of examples and illustrations.

The scope of this work is limited to data structures used in Ethereum. We do not discuss in detail areas such as block mining, verification, execution, Smart contracts creations, etc. However, in some cases we mention these areas as well to support explanation of the data structures. 

This paper is structured as follows. Fundamental data structures are described first, extension of these structures in the form of Merkle Patricia Trie is formulated then. After that the structure of block is povided, followed by detail about Trie structures used for Ethereum data representation.  

\section{Background: Trees and Tries}

Blockchain platforms operate with enormous number of keys that must be looked-up from underlying data sources. Due to distribute nature, they furthermore need to be trustworthy. For these reasons, blockchain platforms try to employ effective algorithms for searching and cryptography. They mostly use and extend structures mentioned in following sections.

\subsection{Search Trees}

One of the standard methods for effective searching are trees. They come in various forms, some of them described by Knuth in his famous book The Art of Computer Programming,  Volume 3 \cite{Knuth1998art3}, where very a common binary search tree is described in Section 6.2.2.

The binary search trees may form a prefix tree where keys with the same prefixes share the same path from the root to the leafs. A key is searched in such a tree so that the key is iterated from the first to the last character and for every character it is checked if there is a child node in the tree that match the character. The tree traversal descent to the child node, next character is picked-up and the same step repeats. 

Lets us detail the algorithm on a binary encoding of the tree. In this case every node may branch to left or right depending on binary zero or one. All possible valid keys may be stored in such a key in its binary form. The nodes are composed from the root to children to represent all the keys stored in this tree, for every node branching left or right to encode binary one and zero for current position in one of the keys.   Fig. \ref{img:merkle-radix-tree}, the left hand side tree, shows an example of such a prefix tree. It  encodes four binary keys (shown as the legend). These keys form four paths from the root node to leafs (coloured nodes). In contrary to keys in the example, alternative keys such as $100$, $101$ are not encoded in this tree as there is no path from root to leaves representing them. 

This data structure has a benefit of memory saving for data where keys have frequently the same prefixes. Example use-cases me be storing of IP4 addresses as such addresses have the same prefixes for the same IP segments. For example an IP4 CIDR block $192.168.56.0/30$ contains a 30-bit value which is always the same and starts to differ only for remaining 2-bits. In this case, it is effective to store all possible hosts of such a network (actually only four possible hosts), in a prefix tree. The prefix tree will need $30 + 2^2=34$ bits, while storing every address independently will need $32 \times 2^2=128$ bits. 

\subsection{Patricia Trie}

\begin{figure*}[ht]
\includegraphics[width=15cm]{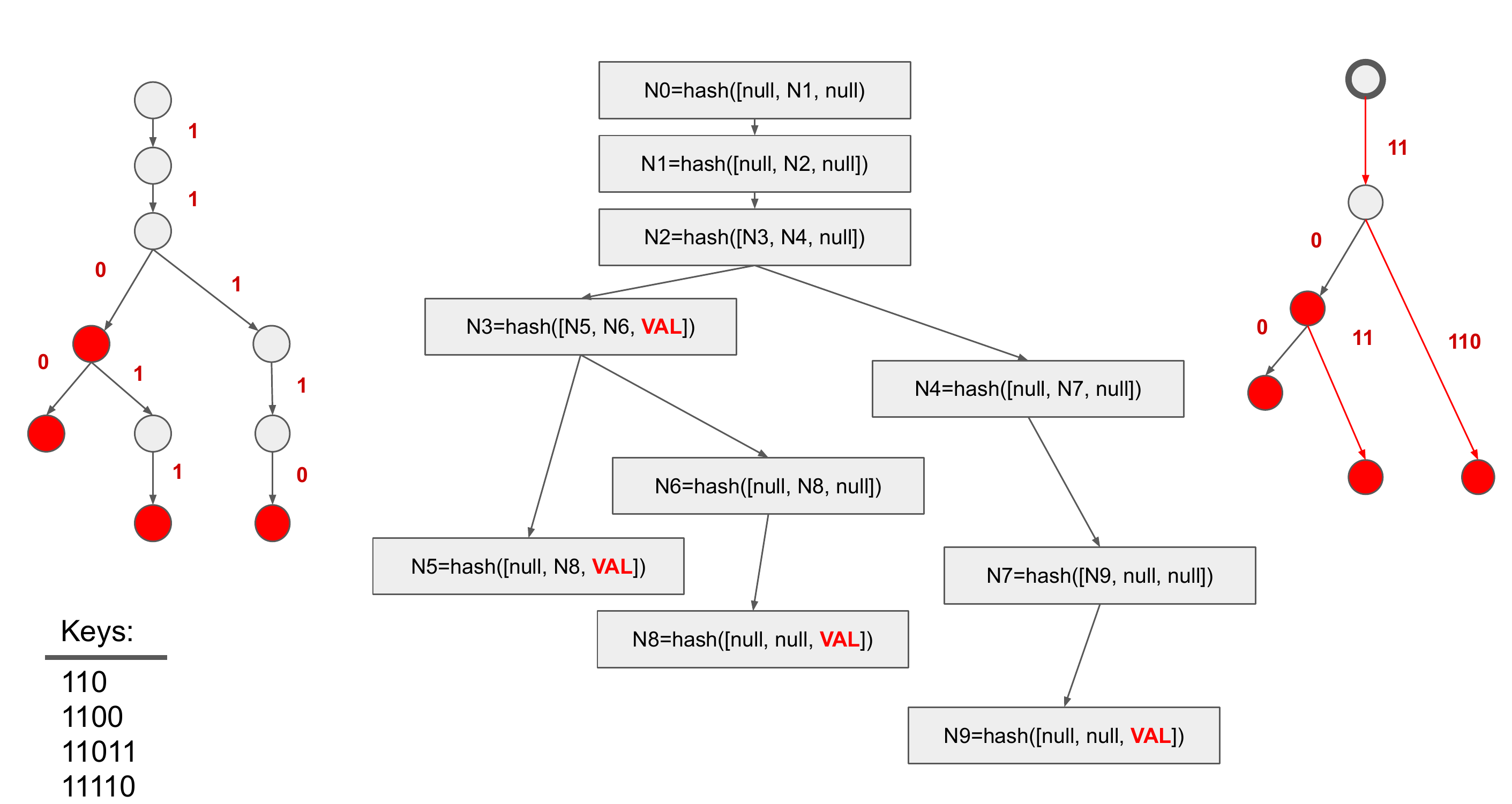}
\caption{Prefix Tree, Merkle Tree, Patricia Trie}
\label{img:merkle-radix-tree}
\end{figure*}

Obvious drawback of a search tree is that it must descent through several children in the cases there is no branching (i.e. a node has only one child).  For instance, in example from Fig. \ref{img:merkle-radix-tree}, two nodes must be crawled to reconstruct all keys prefixed with `$11$'.
Although the example shows a simple binary tree, the same problem persists for N-ary trees. There may be always situations where the only branch exists. The problem increases when the keys have long common parts as it creates long single-branch paths in the tree. 

To solve this problem, Fredkin proposed Trie Memory in his paper \cite{Fredkin1960trie}. He calls the new structure Trie because it serves for information ret\textit{rie}val. He defines data representation in a form of a table where columns of a table are nodes and rows are branches. The number of rows depends on the number of branches. For instance, to represent binary data, two rows are needed.  In contrast,  to represent hexadecimal  strings, it needs 0 to F rows. The number of columns varies depending on the number of branching. If the same data are represented in a tree, the depth of the tree is  equivalent to the number of columns.

Every cell of this table can contain either a reference to a child node or a string. The reference to a child is represented by a number of a column to jump-to.  If the string is stored, it is then equivalent to a suffix of the key.  It also terminates a current path. 

The look-up starts that the input key is traversed by characters. The character at a current position is found in a row first. If this row contains a string, this path terminates and a key is found.  If the cell contains a link to another node (i.e. another column), this column contains all possible branches of all encoded keys. To find a particular key, the next character from the input key is taken and the row matching this character is found. This row together with current column determines a table cell. This cell is again analysed and the algorithm recurses. In a case the cell is empty (i.e. it contains neither next column nor the string), the input key is not encoded in the table.

\begin{figure}[ht]
\includegraphics[width=8cm]{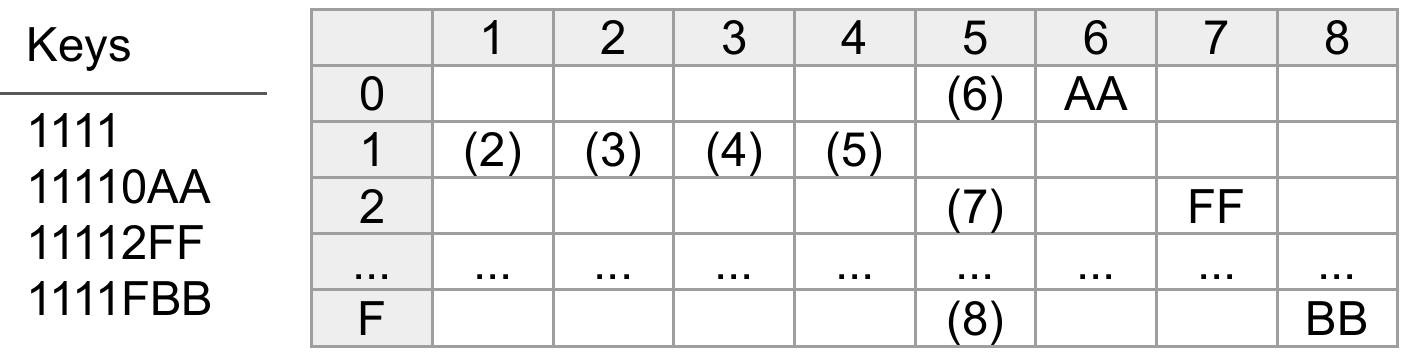}
\caption{Memory Trie Example}
\label{img:memory-trie}
\end{figure}

Figure \ref{img:memory-trie} shows an example of memory trie where four keys with hexadecimal characters are encoded. First four characters (a number `1') are the same for all keys. For this reason, the first four columns only refers to the next column. Then the branching happens, i.e. the column labeled `5' contains references to columns `6', `7', and `8'. The column `5' itself denotes next character in the keys on respective rows -- rows `0', `2' and `F' corresponding to fifth characters in all encoded keys. Columns `6,7,8' finally contain suffixes of the keys as no more branching is necessary.

This data structure is effective in terms of speed because it can be represented as a two-dimensional array and thus accessed via indexes in this array. On the other hand, it has big memory overhead especially when the structure is sparse. This problem may be overcame by representing paths by linked lists as proposed by Briandais \cite{Briandais1959file} so that all possible branches from one node are represented as a tree, composing a forest of tries from all nodes.  

Another drawback of original Trie structure is that it groups only common suffixes of the keys. In other worlds it speed-ups look-up of keys that ends with the same suffixes. Of course, datsets with keys having rather the same prefixes can be modelled as well simply by saving the data in reverse order. However, this  does not solve situation when keys tend to have common paths (in the middle of keys) in general. 

This drawback was addresses by Morrison in his work called Practical Algorithm To Retrieve Information Coded in Alphanumeric \cite{Morrison1968patricia}, i.e. Patricia. This structure is a trie using one more optimisation. Every node contains a number of bits that can be skipped when matching a key as all the keys in the dataset have this path the same. When this amount of bits is skipped, the tree branches left or right based on next bit.  

An example in Fig. \ref{img:merkle-radix-tree} visualises a Patricia Trie on the right hand side. All the keys in the dataset starts with $11$ and thus two bits may be skipped (we visualise this as the edge labelled $11$ outgoing from the root node). So the branching starts at the third bit of the keys. Keys suffixes $11$ and $110$ are also grouped, but this is already a feature of original Trie design, we only visualise it here as a tree instead of a table. 

\begin{figure}[ht]
\includegraphics[width=8cm]{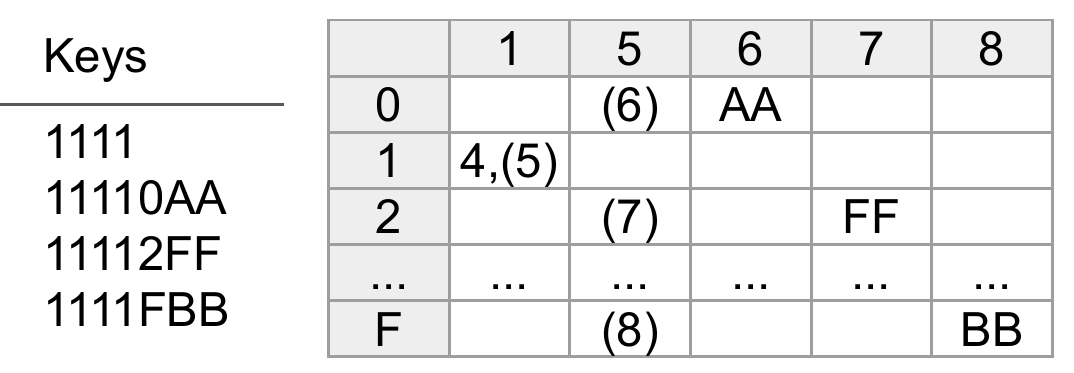}
\caption{Memory Patricia Trie Example}
\label{img:memory-patricia-trie}
\end{figure}

Let us note that Patricia Trie was originally proposed for binary trees, but it is practically used nowadays for N-ary trees. It is for instance easily possible to represent example from Fig. \ref{img:memory-trie} in a Patricia format. It is visualised in Fig. \ref{img:memory-patricia-trie} where the tuple `4(5)' says that before accessing column `5', four columns (i.e. the same characters) have been skipped.  

\subsection{Merkle Trees}
Ralph Merkle developed and described a data structure for digitally signing datasets with desire for fast verification of data consistency. He called it a Merkle Tree \cite{Merkle1987tree}. The Merkle Tree computes cryptography hashes so that every node in the tree contains a hash of its children. The leafs of the tree links  values from the encoded datasets. Naturally, an asymmetric hash is used not to be able to recover original values. 

The purpose of this data structure is to allow parties to verify consistency of a data set without the need to exchange the dataset itself. It is especially useful in distributed environments, where consistency of replicated data must be verified while analysis of physical data is impractical due to their size. 

Merkle Trees, sometimes called Merkle proofs, prevent malicious or unintentional modifications of the data by providing a unique hash that identifies the data set. Another data set is provably  the same only if it computes to the same hash, otherwise the data set is different. Due to this feature, participants in any data exchange may easily verify that they are getting the data they expect, or participants synchronizing data sets may easily verify that they both ended up with the same results. 

Since every subset of the data is hashed as well, the participants may exchange only sub hashes if they do not need to work with complete datasets. In a case a value from any node is manipulated, all the hashes from this node up to the root do not match, and this error is reveled without the need to traverse complete dataset. 

Blocchain platforms in nature involve distributed parties that do not trust, while they need to manage common data. In this setting, Merkle trees server as ideal solution as they provide a proof of validity in term of hashes. In addition, the hashes are small and thus make it ideal for transfer over the internet. 

Fig \ref{img:data-structure}, the tree in the middle, illustrates this principle on a binary tree. As it may be seen, every node from bottom-up is hashed, visualised in abstract terms by a function $hash$. The leaf nodes hash only themselves, but all remaining parents hash recursively hashes of their children. When the data consistency must be verified among parties, a hash of any node can be verified. 

The Merkle tree originally contains only the hashes while the data set values are stored externally. It means the tree is build on top of a data set, which can be modeled by any alternative data structures. Blockchains and Ethereum in particular proposes a structure that wraps all the hashes and the data in one tree. Its description follows in the next section. 

\section{Merkle Patricia Trie}

Merkle Patricia Trie \cite[Appendix D]{wood2019ethereum} is a data-structure that combines and extends Patricia trie and Merkle tree. The same way as Patricia is designed, it effectively stores primary keys while grouping their common paths in one node. As a tamper-proof data validation, it uses the Merkle Tree. This new data structures is unique in its combining these two data structures in one. It is organized as a tree with every node hashed in the sense of Merkle proof and it groups common paths in the sense of Patricia Trie. 

The Ethereum authors proposed Merkle Patricia Trie with data persistence in mind. This is obvious from reading the Yellow Paper. Although, it reads in Appendix D, paragraph D1: `no explicit assumptions are made concerning what data is stored', it continues in the next paragraph: `implementation will maintain a database of nodes'. Since they expected the implementation to store the tree nodes, it explains the reason why all the keys, the data and the hashes are all stored in one tree structure. In other worlds, the Merkle Patricia Trie represents not only a data structure, but also a database schema.

The design of tightly coupling the data structure with the database scheme made it easy to store the structured data in non-structured databases, such as key-value databases. This intention is evident e.g. in both major clients: Parity and Geth, which both use the key-vale storages, RocksDB\cite{borthakur2013under} and LevelDB\cite{leveldb2014fast} respectively. On the other hand, such designs are usually difficult to optimise. It has been already discussed by researchers that Ethereum is riddled with storage bloat \cite{sanguna2018proposals,kim2019ethanos,patsonakis2019alternative} and also the performance goes down due to read and write amplification caused by continuous computation of hashes \cite{raju2018mlsm}.

This new structure allows for grouping common parts of the keys and suffixes in one node, and it uses nodes for branching only when branching is necessary. These ideas are reused from Patricia Trie, however, this new data structure allows for 16 branches and stores hexadecimal strings instead of bits. When keys have common parts, this part is, moreover, directly stored in the node, instead of storing only the number of bits to skip, as it was originally invented for Patricia. To employ Merkle proof, when the Merkle Patricia Trie refers to a child node, it refers to its hash. 

The Merkle Patricia Trie defines three types of nodes with following structure: 

\begin{itemize}
    \item \textbf{Branch} -- a 17-item node $[i_0, i_1, ... , i_{15}, value]$
    \item \textbf{Extension} -- A 2-item node $[path, value]$
    \item \textbf{Leaf} -- A 2-item node $[path, value]$
\end{itemize}

The \textbf{Branch} node is used where branching takes place, i.e. keys at certain character position starts to differ. First $16$ items are used for branching, which means that this node allows for 16 possible branches for $0-F$ hex character. This concept is familiar from the Memory Trie structures, where branching was constructed as a column in a table. The $i-$position contains a link to a child node whenever the child exists, i.e. this position corresponds with a next character (hex $0-F$) in a key. The $17$th item stores a value and it is used only if this node is terminating (for a certain key).

The \textbf{Extension} node is where the compression takes place. Whenever there is a part common for multiple keys, this part is stored in this node. In other words, it prevents descending several times should it follow only one path. This node, in other words, resembles a Patricia feature of number of bits to skip.   

The \textbf{Leaf} node terminates the path in the tree. It also uses a compression as it groups common suffixes for keys in the $path$, which is a concept re-used again from Memory Trie. 

\begin{figure}[ht]
\includegraphics[width=8cm]{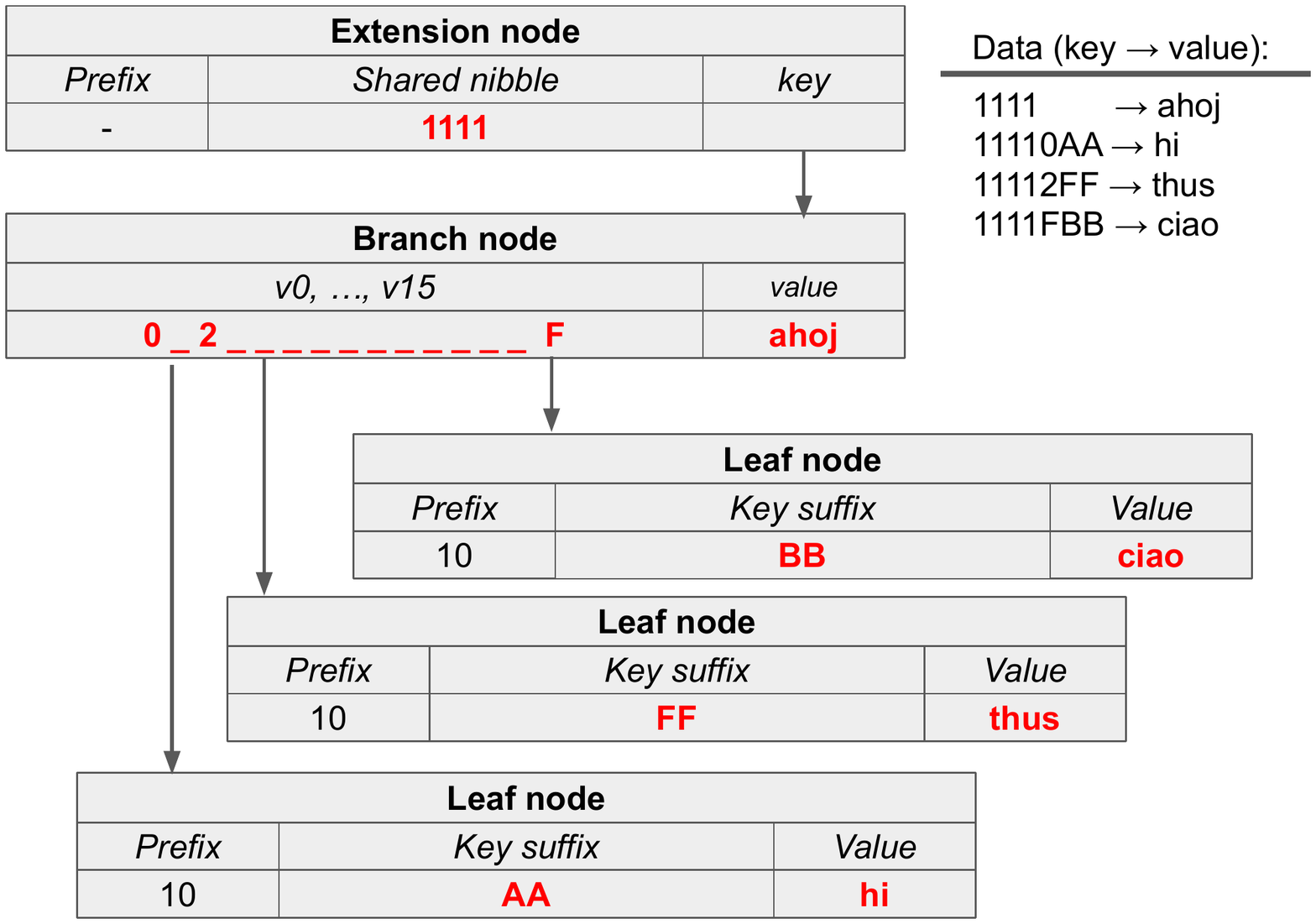}
\caption{Merkle Patricia Trie}
\label{img:merkle-patricia-plain}
\end{figure}

An example of Merkle-Patricia Trie is illustrated in Fig. \ref{img:merkle-patricia-plain}. In contrast to the previous example from Fig. \ref{img:merkle-radix-tree}, this example shows keys as hexadecimal numbers. As it may be seen in the example, all the keys have the same prefix $1111$ and for this reason the extension node is created as a root -- it groups this prefix. Then the keys start to differ, which is captured by creating the branch node. This node branches for characters at the same position of all keys. In particular, branches are created for characters `$0$', `$2$', and  `$F$'. The key having solely `$1111$' terminates here, which means that the value for this key is stored in this branch node. Rest of the keys form leaf nodes as there is no more branching. The leaf nodes store suffixes of each key and store values for the keys as well.

This example does not detail computation of hashes and concentrates only on `Patricia part' of this data structure to better visualise nodes forms. In reality, every link from a parent (i.e. arrows in the image) must refer to a hash representation of the child. This concept has been already shown in Fig. \ref{img:merkle-radix-tree} in the Merkle Tree. We can on a high level write here that a link to a child node is $hash(node)$. However, the Merkle Patricia Trie structures nodes are multi-element (structured) records while hash functions compute a hash out of binary strings. For this reason, the nodes must be encoded from a structured form to a byte array. This will be detailed in the next section. 

\section{Data Structures Encoding}

This section details encoding of trie structures, which allows for representing structured format of trie nodes as byte arrays. This makes the nodes suitable for hashing and persisting. Two main steps are taken to convert a structured node into an array: Hex Prefix Encoding and Recursive Length Prefix. 

\subsection{Hex Prefix Encoding} 

Data stored in Merkle-Patricia trie are encoded to store them as a sequence of bytes. Leaf and Extension nodes are encoded using a \textbf{HP} Hex-Prefix-Encoding function. HP is formally defined in Yellow Paper \cite[Appendix C]{wood2019ethereum} and its purpose is to encode any sequence of nibbles into binary stream so that two (4-bit) nibbles compose one (8-bit) byte. 

More formally, nibbles are encoded as a sequence: $\{16 x_i + x_{i+1} | i \in [0, \dots, ||x||-1)\}$

In addition, this sequence is prefixed by a nibble, which has two purposes. Second-lowest bit is a flag saying if the node is either a Leaf or an Extension. This bit is set `$0$' for Extension and `$1$' for Leaf node. The lowest bit of the same nibble is a flag set to `$0$' for even length of the encoded data and `$1$' for odd length.  This is depicted in the table:

\begin{center}
\begin{tabular}{ r r  c c }
Hex  &  Bits    &    Node Type     &   Path Length \\
\hline
   0    &     0000    &   Extension    &    even \\       
   1    &     0001    &   Extension    &    odd    \\     
   2    &     0010    &   Leaf          &   even \\        
   3    &     0011    &   Leaf          &   odd  \\
\end{tabular}
\end{center}

The bit saying if the number of nibbles is odd or even is important for padding to be able to reconstruct original values for decoding. The number of persisted nibbles must be always even, because the resulting byte array will be always stored to an even length (because standard computers cannot store only a half of byte). For example, if an even length was not assured and a hex value `$0x10$' was encoded it would not be  possible to determine if original nibbles were two: with values `$1$' and `$0$', or the original nibble was one: with value `$1$'. On the other, if it is given that the number of nibbles is always even, the value `$0x10$' can represent only two values: `$1$' and `$0$'. 

Following this first prefix nibble, another nibble with value `$0$' is inserted if the number of nibbles is odd. This zero is a padding to extend the number of nibbles to an even number. If the number of nibbles is even already, this additional nibble is not used and the encoded path continues immediately after the first nibble.

Let us summarise the HP in following examples. Brackets mark one byte with two nibbles. As it may be seen, a padding zero has been added as a second nibble for even length, while the data follows directly the first nibble for odd length:

\begin{center}
\begin{tabular}{ r l  c c }
Prefix  &  Payload    &    Node Type     &   Path Length \\
\hline
     $[0, \mathbf{0}]$ & $[x_1 x_2][x_3 x_4]$ & Extension & Even \\
     $[1, x_1]$ & $[x_2 x_3][x_4 x_5]$ & Extension & Odd \\
     $[2, \mathbf{0}]$ & $[x_1 x_2][x_3 x_4]$ & Leaf & Even \\
     $[3, x_1]$ & $[x_2 x_3][x_4,x_5]$ & Leaf & Odd \\
\end{tabular}
\end{center}

Let us sum-up with two examples showing one even and one odd strings encoded via \textbf{HP}. The first example encodes odd number of characters. For this reason the prefix `1' (could be `3' as well) follows immediately with the payload, two characters stored at one byte. The second example contains even number of character. For this reason the prefix `0' (or possibly `2') is followed by additional padding `0' and then by the payload. 
\begin{equation*}
\begin{split}
[5, 6, 7, 8, 9] &\rightarrow [15, 67, 89] \vee [35, 67, 89]\\
[4, 5, 6, 7, 8, 9] &\rightarrow [00, 45, 67, 89] \vee [20, 45, 67, 89]
\end{split}
\end{equation*}

\subsection{Recursive Length Prefix}
After HP is used, it encodes node structures into compressed byte arrays. However, trie nodes still contain more arrays in every node  (i.e. Extension node is 17-item structure -- thus contains 17 values/arrays, Leaf and Extension contain two arrays). It is practical to flatten these arrays to one byte array structure for later persistence in a key-value storage. For computing the hash, it is necessary to flatten the arrays as the hash function cannot be applied to structured forms.  

Yellow Paper \cite[Appendix B]{wood2019ethereum} defines a \textbf{RLP} function, which serialises a set of input arrays into one array. Hence, using this function a node from the trie may be converted a into flat byte array and persistent into a key-value storage as one value.  Yellow Paper details the RLP function formally, but in essence this function serialises the arrays so that it outputs all sub-arrays in one long array. Every sequence representing original sub-array is prefixed with its length and a bit-mask to determine if the sub-array was originally an array or a string. Every level of flattened structure is also prefixed by total length of all sub-arrays.  Details of RLP encoding follows: 

\paragraph{Byte Values Up-to 127} 
Such single bytes are stored directly without modifications. 
We can simply write that input byte $b$ is transformed:
\begin{equation}
RLP: b \rightarrow b
\end{equation}

\paragraph{Strings With Length Up-To 55 Characters}  
Transformation used for encoding input Strings $s$ is:
\begin{equation}
RLP: s \rightarrow [\mathrm{0x80} + ||s||, s] 
\end{equation}
 where value $||s||$ donates length of strings $s$, the string itself follows this prefix. As it may be seen, transformation of such short strings (up to 55 characters) is very efficient, as it needs only one additional byte, which stores the prefix together with the string length.

\paragraph{Arrays with Total Length Up-to 55 Characters}
For arrays with accumulated length up-to $55$ characters, including possible nested arrays, the RLP function transforms a set of arrays $S$ as follows:
\begin{equation}
RLP: S \rightarrow [\mathrm{0xc0} + ||ex(S)||, ex(S)]  \\
\end{equation}
The function $ex$ expands the set $S$ so that it computes RLP encoding on every array from this set:
\begin{equation}
ex : S \rightarrow (RLP(s_i)|  s_i \in S)
\end{equation}

As it may be seen, encoding of such arrays is again efficient needing only one byte to store the prefix together with total length. 

\paragraph{Strings With Length Up-To $\mathrm{2^{64}}$ Characters}
For string longer then 55 characters, the RLP function for String $s$ is computed as follows: 
\begin{equation}
RLP: s \rightarrow [\mathrm{0xb7} + numB(||s||), ||s||, s]    
\end{equation}
The function $numB : N \rightarrow N$ computes number of bytes required to store an input value. In other words, having length of the string $s$, this functions says how many bytes are required to store this length. For instance strings up-to $\mathrm{0xFF}$ characters need one byte to represent their length, lengths between $\mathrm{0xFF}$ and $\mathrm{0xFFFF}$ need two bytes, etc. Furthermore, this variant of RLP produces length of $s$s, followed finally by the string itself.  

This variant of RLP function requires more bytes to represent the prefix. Unless the strings are very large, this function is still reasonable efficient. For instance, strings up-to the length $\mathrm{0xFFFF}$ require three extra bytes to encode (1 byte for the prefix, 2 bytes for the length). 

\paragraph{Arrays With Length Up-To $\mathrm{2^{64}}$ Characters}
For nested arrays with accumulated length higher then $55$ characters, similarly as in the previous cases, the RLP function prints a prefix, a number of bytes required to represent the length, the length itself and recursively RLP encoded payload follows. Formally arrays $S$ are encoded:
\begin{equation}
RLP : S \rightarrow [\mathrm{0xf7} + numB(||ex(S)||), ||ex(S)||, ex(S)]
\end{equation}

\paragraph{Payloads With Length Higher Than $2^{64}$}
Neither strings nor arrays of this size can be encoded.

Let us illustrate the RLP function on four examples. The first example encodes plain string $ABCD$ -- it uses the prefix $\mathrm{0x80}$ plus size of the string: $4$: 

\begin{equation*}
\begin{split}
ABCD \rightarrow &[\mathrm{0x80} + 4, A, B, C, D]	
\end{split}
\end{equation*}

The second example encodes two arrays with $AB$ and $CDE$ -- the prefix $\mathrm{0xc0}$ is used plus length of remaining encoded payload: 

\begin{equation*}
\begin{split}
[AB][CDE] \rightarrow &[\mathrm{0xc0} + 7, \mathrm{0x80} + 2, A, B,  \mathrm{0x80} + 3, C, D, E]	
\end{split}
\end{equation*}

Third example encodes a long string. Let us assume we have a string with the length of 300 characters, to save space we shortcut this string and write only: $A \cdots B$. To express the length, we need in hexadecimal form $\mathrm{0x12C} \equiv 300$. This number requires two bytes. We use the prefix for long strings: $\mathrm{0xb7}$ and add $2$, the two bytes representing the length are stored next, and the string itself follows: 

\begin{equation*}
\begin{split}
A \cdots B \rightarrow &[\mathrm{0xb7 + 2, 0x1, 0x2C}, A, \cdots, B]	
\end{split}
\end{equation*}

The last example recursively stores two arrays $[A \cdots B][ABCD]$, first contains a long string while the other one a short string. We encode bottom-up, i.e. we apply two rules for long and short strings used in previous examples. It creates two byte sequences $[\mathrm{0xb7 + 2, 0x1, 0x2C}, A, \cdots, B]$ and $[\mathrm{0x80} + 4, A, B, C, D]$ already known from previous examples.  To finish encoding, we use the prefix for long arrays first: $\mathrm{0xf7}$. Then we compute the lenght of the new payload, which is $\mathrm{0x134} \equiv 308$. It still requires two bytes to store. The final encoding will contain the prefix $\mathrm{0xf7}$ plus 2, the bytes $\mathrm{0x1}, \mathrm{0x34}$ and finally two already encoded arrays, in total: 

\begin{equation*}
\begin{split}
[A \cdots B][ABCD] \rightarrow &[\mathrm{0xf7 + 2, 0x1, 0x34},  \\& \mathrm{0xb7 + 2, 0x1, 0x2C}, A, \cdots, B, \\& \mathrm{0x80} + 4, A, B, C, D]
\end{split}
\end{equation*}

The purpose of RLP function is to convert a structured node form into a single byte array. For JavaScript and web-developers this concept is familiar as a variant of JSON. However, RLP is designed as more memory efficient because it needs only a couple of prefix bytes. In contrast, JSON is a text-markup that includes special characters to mark beginning and end of structures (i.e. arrays, objects, etc). For instance, JSON always needs at least two bytes to mark beginning and end of an array, using `$[$' and `$]$' characters. On the other hand, due to repeated characters, JSON can be often widely compressed \cite{afanasev2017performance}.

The RLP function is also prerequisite for hashing the nodes. Once the node is turned into a byte array, an asymmetric hash function can be applied. This fundamentally enables the Merkle proof in Merkle Patricia Trie. 

\begin{figure}[ht]
\includegraphics[width=8cm]{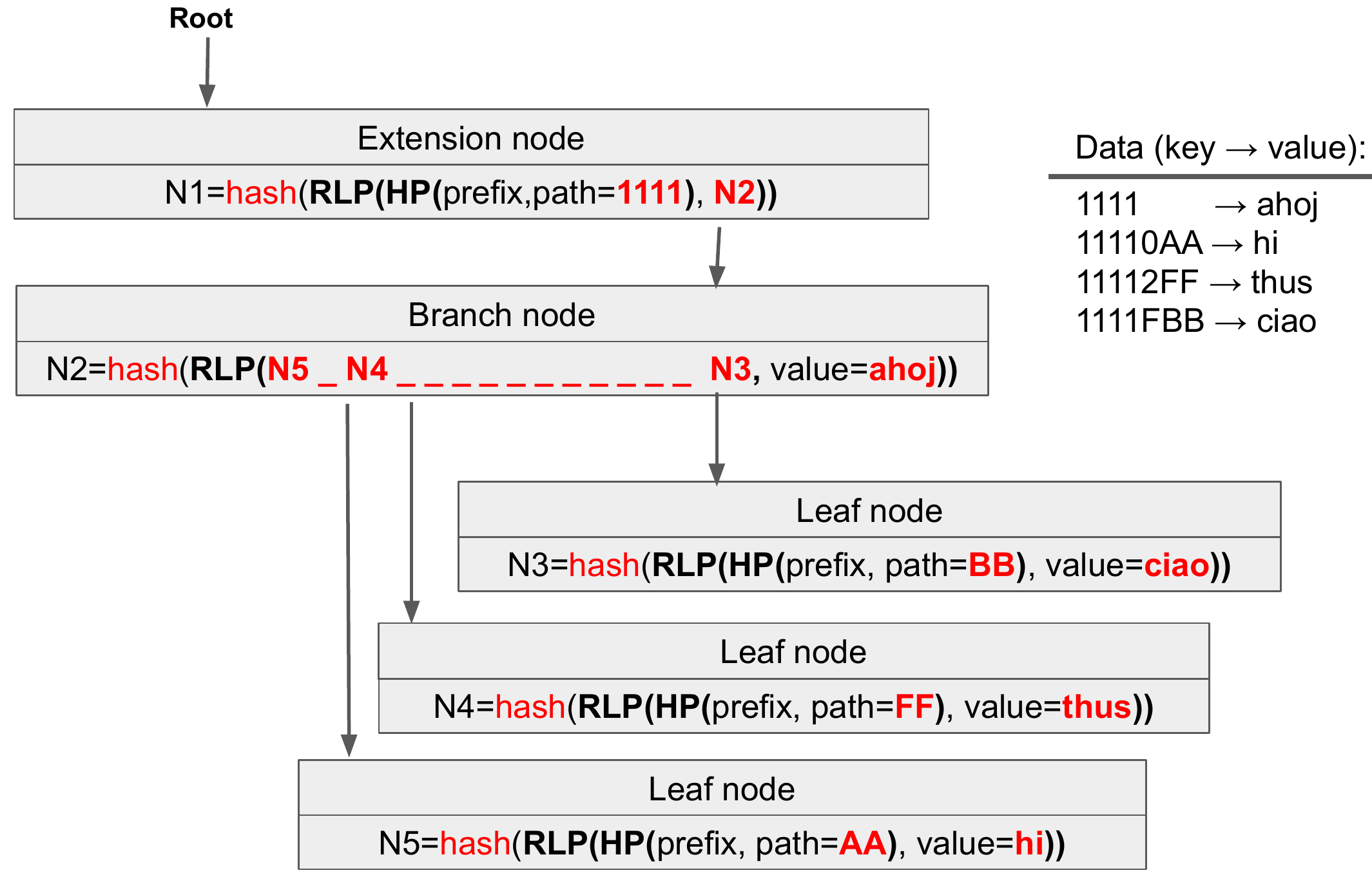}
\caption{Ethereum Encoded Merkle Patricia Trie}
\label{img:merkle-patricia-encoded}
\end{figure}

The example in Fig. \ref{img:merkle-patricia-encoded} complements previous Fig. \ref{img:merkle-patricia-plain} by visualising encoding of nodes, which is used in the Ethereum. The image shows that paths in Extension and Leaf nodes are encoded by HP first, then each node is encoded into a byte array by the RLP encoding. Then, the whole node can be hashed. After that, every parent node refers to its children via this hash adding the Merkle proof to every node. 

\subsection{Persistence in Key-Value Storage}

Once nodes in the trie are encoded by HP and RLP, every node is then represented only as an array of bytes. It makes it suitable for storing in a key-value database as such databases are typically designed for representing keys and values as simple data structures, i.e. byte arrays. 

Since all the nodes are hashed, it is not possible to reveal their original value. For this reason, it is not possible to store only the hashes in the database. Ethereum solves this by storing both hashed and plain (only RLP encoded) values together in the database. Ethereum uses a hash function Keccak \cite{Bertoni2011Keccak}, which is an original implementation of SHA3 available before SHA3 became official.  

Every node is stored in the database as a $key \rightarrow value$ pair, where the key is a hashed RLP representation of the node and the value is an RLP encoded node. We can write that one node is represented in the database as: 

\begin{equation}
    keccak(RLP(node)) \rightarrow RLP(node)
\end{equation}

In a key-value storage format: $key \equiv keccak(RLP(node))$ and $value \equiv RLP(node)$. Content of every node differs based on the node type:
\begin{itemize}
    \item \textbf{Extension} $node \equiv [HP(prefix + path), key]$
    \item \textbf{Branch} $node \equiv [branches, value]$
    \item \textbf{Leaf} $node \equiv [HP(prefix+path), value]$
\end{itemize}
where $key$ in the Extension node is a hashed key of a child node. In the Branch node $branches = (key_i | 0 \le i \le 15)$ and $key_i$ is a hashed key of a child node if the branch exists, otherwise is empty. For both node types, the key is computed by an already mentioned formula: $key = keccak(RLP(node_c))$ and $node_c$ is a child node.

Let us note that the hash representation for the key is computed from the whole node, based on its RLP representation. Since the node comprises hashes of child nodes (unless it is a leaf), this essentially build the Merkle part of this data structure as a hash of a current node is computed on top of child hashes. 

Finally, Ethereum uses one more optimisation: node inlining. If a RLP encoding of a child node is shorter then 32 bytes, no hash is computed and this node is stored directly inside a parent node. This technique allows for furthermore compacting the data structure for short nodes. 

\section{Blockchain, Blocks and Transaction}

Every Blockchain consists of a sequence of blocks, which are bound together via cryptography hashes. A block contains a body and a header. The header holds a meta-data about this block.  One of the most important values stored in the header is the hash of the previous block, which composes the chain. The purpose of every block is to execute transactions. Stakeholders participating in a blockchain ecosystem submit new transactions, which represent usually their desire to transfer funds between user accounts. Once the transactions are validated, the transactions are wrapped in a block, the block is attached to the blockchain via the hash and it becomes part of the chain forever. 

The overall idea of blocchain is generic and particular blockchain protocols may store data such as blocks and transactions differently. In the case of Ethereum, the universal data structure used is Merkle Patricia Trie. All the information flowing through the blockchain are modeled using instances of this structure. 

The transactions are encoded in Transaction Trie furthermore detailed in Section \ref{sec:transaction-trie}. A transaction represents an atomic operation, which is recorded in this trie, and may be investigated any time later. In addition, this operation changes a state, i.e. influences other participants in the network. Most typically, the transaction causes that one account's balance is deducted while the other account's balance is increased, i.e. the funds flow from one user to another one. These always rolling state changes are reflected in World State Trie, described in Section \ref{sec:world-state-trie}.  Consequence of this design is, that every block contains in its header links to various Merkle Patricia Tries representing transactions and current state. 

Moreover, the block in Ethereum contains a list of ommers. This list consists of headers of blocks that were successfully mined, but they eventually did not became part of the mainline. Detail description of mining process is out of scope of this paper. Apart from the Yellow paper a readable explanation is provided e.g. by Dannen \cite[Chapter 6]{Dannen2017introducing}. In  brief, the process is as follows: blockchain stakeholders propose their transactions into the network first. A group of specially dedicated stakeholders -- miners -- responsible for validating the network, then pick-up these transactions and wrap them in blocks. To do so, they must verify transactions, compute a hash of a parent block, execute (artificially) resource demanding computation to prove that they invest considerable effort in mining, they get reward for their work and the block became part of the blockchain.  Let us note that the resource demanding computation is part of so the called Proof-of-work consensus protocol, other protocols may work differently as summarised e.g. by Zhang \cite[Section 4]{zhang2019security}.

However, this process may produce more blocks in approximately the same time. These redundant blocks are not thrown, but became ommers, and miners still get a reward, but smaller. Since the tail of the blockchain may contain at any moment these branches, a consensus protocol is used to determine the leading path. Ethereum uses a modified scheme in order to generate consensus based on the protocol called GHOST and proposed by Sompolinsky and Zohar \cite{Sompolinsky2015secure}. This protocol in brief selects a path to which the most computational effort has been invested. This path then becomes a mainline in the blockhain and next blocks are added only to this path. Transactions in the newly added block are finally executed as soon as the consensus on the leading path is made. 

Let us note that miners are not forced to follow the consensus, but their effort on mining blocks outside of the mainline becomes meaningless after a chain of 6 consecutive ommers is created as no reward is paid for longer ommer chains. This number of ommers has been estimated as sufficient for all stakeholders to get information about the consensus being made. Ethereum configures difficulty of mining to propose a new block approximately every 15s, which gives about a minute and half for all miners to get information about new consensus being made. 

We can find that a block $B$ contains a header $H$, transactions $T$ and ommers, sometimes refered to as uncles $U$: 

\begin{equation}
    B = (H, T, U)
\end{equation}

\begin{figure*}[ht]
\includegraphics[width=15cm]{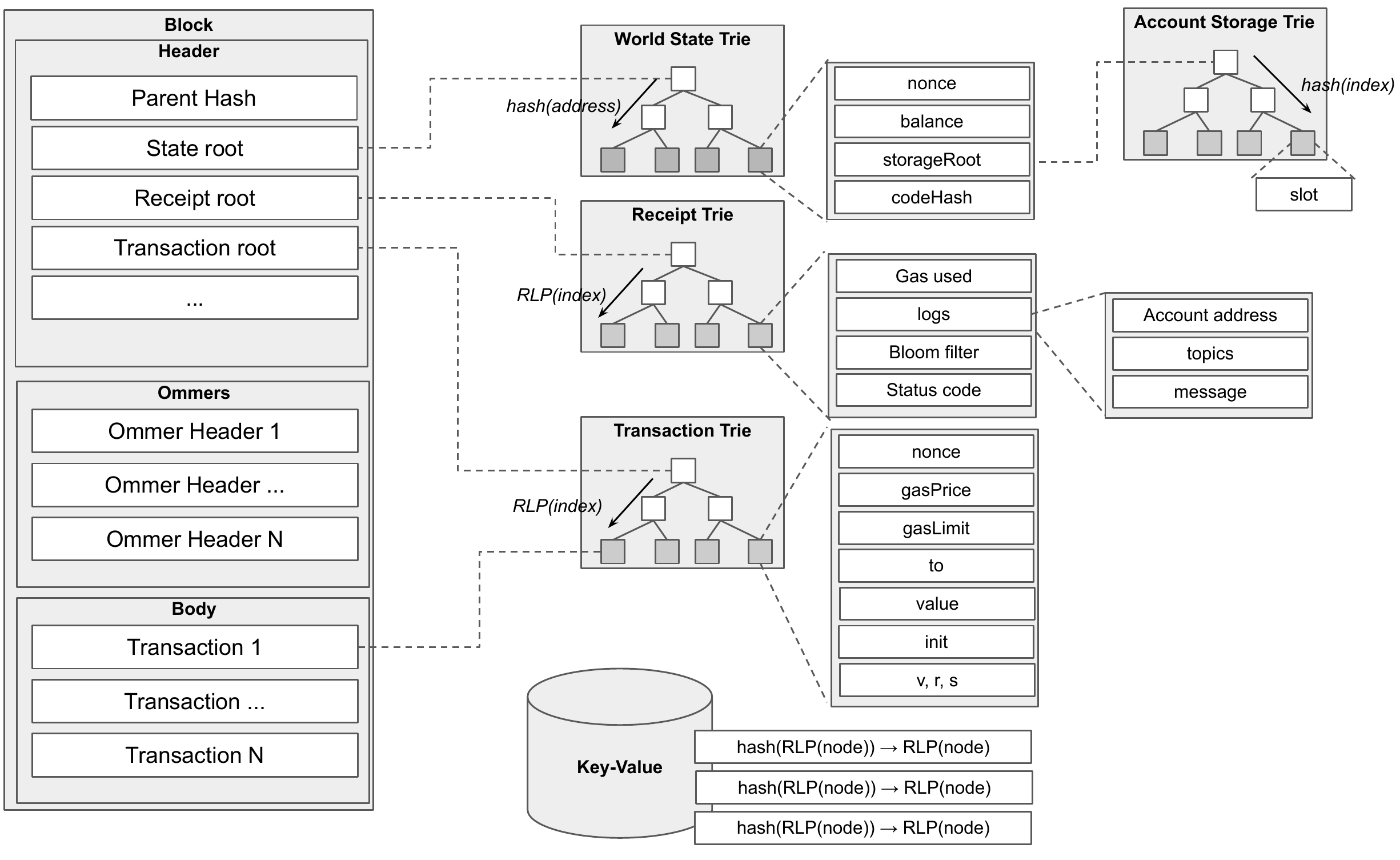}
\caption{Ethereum Data Structures}
\label{img:data-structure}
\end{figure*}

Ethereum blockchain structure is visualised in Fig \ref{img:data-structure}. As it may be seen, the block header links various tries and the body contains the transitions.  All the trie based structures are described in Section \ref{sec:ethereum-structures}. For this reason, this section will detail content of the block header only. The visualisation shows only selection of header fields, the complete list of header $H$ fields is:

\begin{itemize}
    \item \textbf{parentHash} The Keccak 256 bit hash of the parent block. This field connects blocks in a chain.
    \item \textbf{ommersHash} The Keccak 256 bit hash of list of ommers. 
    \item \textbf{beneficiary} An account address of a user who mined this block and receives reward. 
    \item \textbf{stateRoot} The Keccak 256 bit hash of a root of the World State Trie.
    \item \textbf{receiptsRoot} The Keccak 256 bit hash of a root of the Transaction Receipt Trie.
    \item \textbf{transactionsRoot} The Keccak 256 bit hash of a root of the Transaction Trie.
    \item \textbf{logsBloom} A filter relating logs hashes with the log records.
    \item \textbf{difficulty} A scalar value corresponding to an effort that has to be undertaken to mine this block.
    \item \textbf{number} A scalar value equivalent to an ordinal number of this block. Every new block in the chain gets a number increased by one. 
    \item \textbf{gasLimit} A scalar value containing an accumulated gas limit required to process all transactions in this block.
    \item \textbf{gasUsed} A scalar value containing an accumulated real consumed gas to process all transactions in this block.
    \item \textbf{extraData} A free form, 32 bytes or less long byte array, containing any additional data.
    \item \textbf{mixHash} The Keccak 256 bit hash confirming that a sufficient computation has been spent on mining this block.
    \item \textbf{nonce} A 64 bit value. This value combined with mixHash also proves that the computation has been spend for mining this block.
\end{itemize}

Fields \textit{stateRoot},  \textit{receiptsRoot} and  \textit{transactionsRoot} refer to respective trie data structures and are described in other sections, namely \textit{stateRoot} and the underlying World State Trie is provided in Section \ref{sec:world-state-trie}, Receipts and Transactions tries referred via the fields \textit{receiptsRoot} and \textit{transactionsRoot} are provided in Section \ref{sec:transaction-trie}. The same section also details \textit{logsBloom} filter,  \textit{gasLimit} and \textit{gasUsed}.

Values under the keys \textit{mixHash},  \textit{nonce} and \textit{difficulty} are related to block mining and the Proof-of-work consensus. The Proof-of-work main purpose is to enforce a certain computational power is spend on validating a block, independently on number of nodes (i.e. computers) added in the network. This secures the network by disallowing a party with prevailing computation power to rule the network. Ethereum uses the protocol called Ethash\cite{ethereum2020ethahs} for mining and validating that the mining work has been done. This protocol derives \textit{nonce} which is a cryptography hash proportional to \textit{difficulty}. The difficulty is adjusted along the time to make sure that the time frame spent globally on mining blocks remain approximately the same with growing computational power invested in mining. 

The Proof-of-Work consensus is based on solving a cryptography puzzle. In brief, it works as follows. 
The miners manage a dataset of cryptic hashes and their task is to randomly select a sub-set of hashes from this dataset. This sub-set is hashed together and produces \textit{mixHash}. If \textit{mixHash} is lower than  \textit{nonce} the computation is successful, i.e. the miner has found the correct sub-set, and the miner finding this solution is a winner, the miner is rewarded and the block becomes part of the chain.   

\section{Data Representation}
\label{sec:ethereum-structures}

As it has been mentioned, the fundamental process in blockchains is execution of transactions. Ethereum extends on complexity as it contains additional feature: Smart Contracts. Smart Contracts are as well manipulated via transactions (created, executed), but additionally create a global state. Smart contracts are programs that execute a code and store values in global variables. It means that Ethereum must store not only transactions, but also the Smart Contracts and their values. Ethereum does it using Merkle-Patricia-Tries, their structure will be described in next sections. 

Ethereum recognises following data-structures:  Receipts Trie, Transactions Trie, World State Trie and Account Storage Trie.

\subsection{Transactions Trie}
\label{sec:transaction-trie}

Receipts Trie and Transactions Trie record execution of each transaction and are thus non-modifiable  after the transaction is executed. It means that all performed transactions are recorded forever and cannot be undone. It contrasts to the World State Trie, which is continuously updated to hold the most recent state of the blockchain. The World State is updated to reflect either account balance changes or Smart Contract creation, while the record in Transaction and Receipt Tries become verbatim once the transaction is executed. 

A transaction is mapped in the trie so that the key is a transaction $index$ and the value is the transaction $T$. Both the transaction index and the transaction itself are $RLP$ encoded. It compose a key-value pair, stored in the trie: 

%
%

\begin{equation}
RLP(index) \rightarrow RLP(T)
\end{equation}

The structure $T$ consists of following items: 
\begin{itemize}
    \item \textbf{Nonce}: an ordinal number of a transaction. For every new transaction submitted by the same sender, the nonce is increased. This value allows for tracking order of transactions and prevents sending the same transaction twice. 
    \item \textbf{gasPrice}: A value indicating current price of gas. 
    \item \textbf{gasLimit}: A value indicating maximal amount of gas the sender is able to spend on executing this transaction.
    \item \textbf{to}: Address of an account to receive funds, or zero for contract creation.
    \item \textbf{value}: Amount of funds to transfer between external accounts, or initial balance of an account for a new contract.
    \item \textbf{init}: EVM initialization code for new contract creation.
    \item \textbf{data}: Input data for a message call together with the message (i.e. a method) signature.
    \item \textbf{v, r, s}:  Values encoding signature of a sender.
\end{itemize}

Although this document does not primarily detail execution semantics of the blockchain, some overview must be given to better understand these fields. A transaction can represent one of the three kinds: a simple transactions that transfers funds between accounts, Smart Contract creation, and a Message call. 

All the fields mentioned above are used for all types of transactions, but some of them are not filled depending on the context. In particular, the field \textit{init} is filled only for contract creation and in this case the field \textit{to} is empty. The field \textit{data} is filled only for message call. Finally, both \textit{init} and \textit{data} are empty for transactions that only transfer funds. In this case \textit{to} is filled and it contains a recipient address.  

Three types of transactions and the mapping of respective fields is  summarised in table:
\begin{center}
\begin{tabular}{c | c | c  | c } 
 Field & $to$ & $data$ & $init$ \\
 \hline
 External account & recipient address &  &  \\
  \hline
 Contract creation &  &  & byte-code + data \\
  \hline
 Message call & contract address. & signature + data &  \\
  \hline
\end{tabular}
\label{tab:trans-fields}
\end{center}

\paragraph{Transaction Fee}
The \textit{gasPrice} and \textit{gasLimit} are used for computing a fee paid for executing a transaction, as also explained by Kasireddy in her blog \cite{Kasireddy2017howethworks}. Ideally these values should be aligned with fiat money required to run the transaction,  typically invested into real-world commodities such as electricity and also hardware costs. 

When an Etherum Virtual Machine executes a smart contract, it charges certain gas for every instruction and the storage used. This fee is hardcoded, but has been updated in the history by doing a so called hard forks (i.e. an incompatible changes in the platform). A list of recent forks is provide in a web page \cite{etherchain202hardforks}. For instance recent Istanbul hard fork has changed prices of SSLOD and SSTORE \cite{Olszewicz2019hard}, which is an incompatible and controversial change as it changes price of execution.

The \textit{gasPrice} is set by the transaction submitter. When the miner executes the transaction, the gas charged by the Virtual Machine is multiplied by the gas price and it becomes a reward (in Ether) send to the miner's account.  As in other trades, the gas price is correlated with the market as the smart contract creators want to execute their contracts at minimal price, while the miners tend to select best valued contracts, they essentially need to meet somewhere in the middle. 

Since the blockchain is decentralised and anonymous parties run transactions, this mechanism should guarantee that code of underlying Smart Contracts is optimised not to waste computer power. Senders (creators) of a transaction are incentivized to optimise their code to consume minimal gas to save their money. In contrast, all expended gas go to a beneficiary, which is in the current Prof-of-Work scheme the miner, who is motivated to execute and validate transactions, which keeps the blocchain as such alive. 

The value represented by \textit{gasLimit} is set by the transaction creator and expresses maximal gas a sender is willing to pay for executing his or her transaction. This limit is a kind of fuse, which prevents endless running code that would at the end drained the sender's funds. An endless loop would also, at current implementation, block execution of blockchain as such because the transitions are currently executed synchronously. While there is ongoing research into asynchronous execution \cite{Dickerson2017concurrency,Wst2019ACEAA,Anjana2019efficient,zhang2018enabling}, it is not mainstream yet.

The gas limit should not be under estimated by the sender, because the transaction execution is terminated  if it reaches the gas limit prematurely. All the already consumed gas is, however, not refunded as someone (i.e. the miner) has to already spend computer power to execute before reaching out-of-gas. It means that the sender will end up without his or her transaction being accepted in the blockchain, but still having to pay for (partial) execution. 

\paragraph{Contract Creation}

\begin{figure}[ht]
\includegraphics[width=15cm]{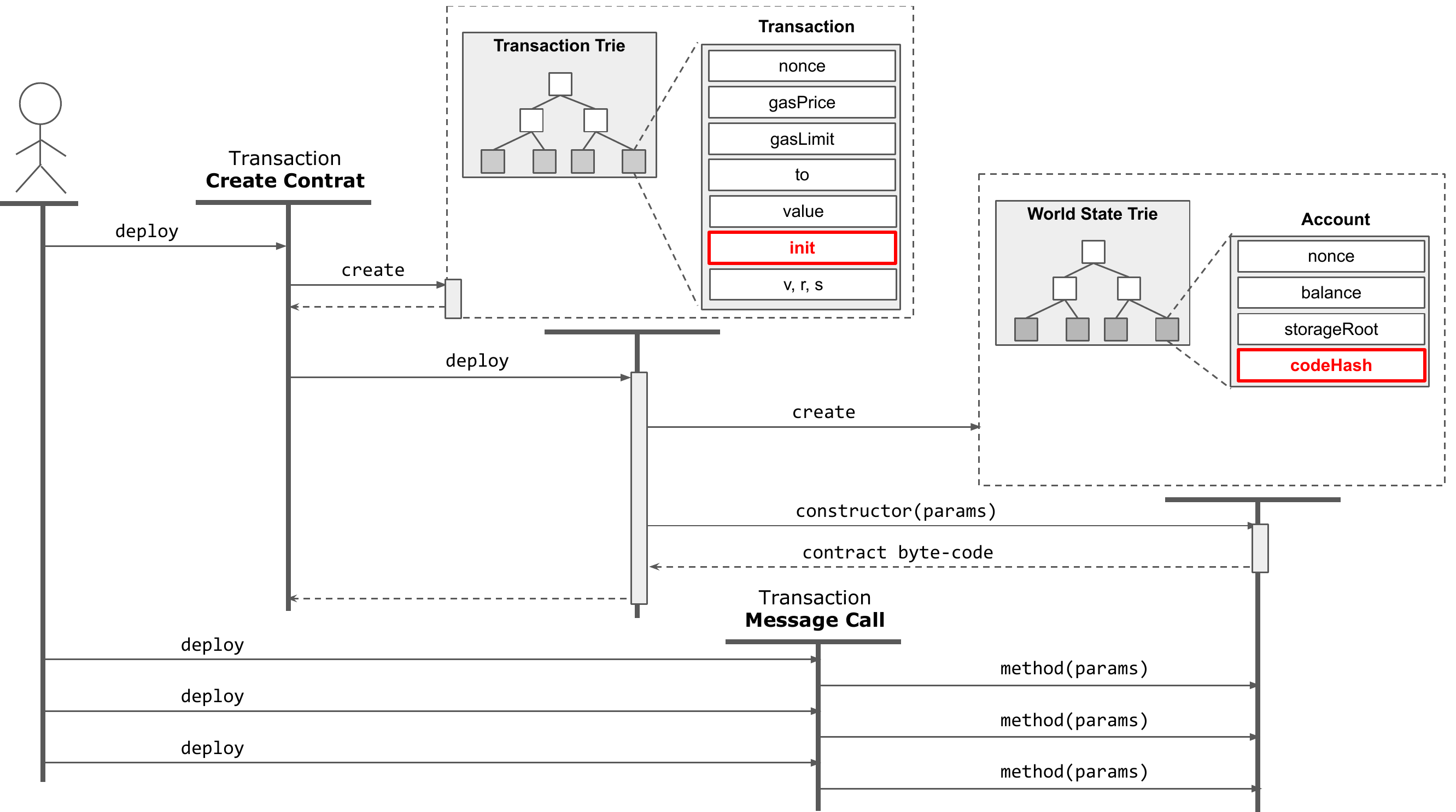}
\caption{Creation of Smart Contract and its Invocation}
\label{img:transaction-create-call}
\end{figure}

A new Smart contract is deployed into the blockchain by a special type of transaction.  The worflow is visualised in Fig \ref{img:transaction-create-call} by the means of sequence diagrams. When a user, visualised as a human actor in the diagram, deploys the transaction, it is executed. Let us note that the transaction is not executed immediately, it must be first validated by a miner and in ad-hoc sequence synchronised by all nodes participating in the blockchain network. This delay is not directly visible from the diagram. 

Once the transaction is executed, it creates a record in the Transaction Trie with all the fields related to the transaction. The diagram visualises it as creating a new actor with the trie (obviously only one item in the trie is created, not the whole trie). For contract creation, the field \textit{init} in this trie is filled with an EVM byte-code that generates a new contract. This byte-code has been previously generated by any smart contract compiler, most typically using a Solidity\cite{solidity2020language} language and its compiler, and arrived with the transaction. 

The filed \textit{to} in the Transaction Trie is empty because part of contract deployment is creation of a new account. For this reason the value for filed \textit{to} is not known yet. A new contract account is created immediately after the smart contract is deployed, showed in the diagram as a creation of a new actor with the World State Trie (again, one item in this trie is created). This new account gets byte-code of the contract stored in the field \textit{codeHash}.  In addition, a new storage represented by a Storage Trie is created and the root of this trie is saved in the field \textit{storageRoot}. Contract creation can already contain initial state (i.e. global variables), which are stored in this trie. 

From the programming language perspective, the initialisation code is implemented as a special method: a constructor, which is executed together with all the instructions in its body. This also means that the constructor can already set-up a state via program variables. The result of the constructor invocation is the contract itself as it returns body of new contract, associated via the account field \textit{codeHash}. Using the object oriented programming terminology, we can say that the constructor creates a contract instance. The constructor can have arguments and in this case they are appended in a binary form to the \textit{init} field. Details about binary format is in a readable form detailed in the Howard's blog \cite{Howard2017contractcreation}. 

The rest of the UML diagram depicts execution of Smart Contract, detailed in next subsection. 

A new contract can also come with initial funds, which are stored in the field \textit{balance}. It means, that not only humans, but contracts as well can hold money. This fact was one of the causes of infamous The DAO attack\cite{Humiston2018thedao}, which allowed an attacker to drain approximately third of that time available Ethers. The DAO was a bidding system, where participants send their bids (i.e. deposits), which were about to be refunded after a certain period. The main attack vector was a security bug in the contract, but such a big theft was possible only because all the money collected from participants were accumulated in one account. 

\paragraph{Message Calls}

Once deployed, contracts can be executed many times. Contract execution basically means execution of its methods, which is triggered by a special transaction, which has a non-empty filed \textit{data}. Such a contract execution is called a message call, which wraps a method signature and data in certain format. This concept is similar to RPC (Remote Procedure Call) or RMI (Remote Method Interface), already well known from other languages such as Java\cite{oracle2020RMI}.

Note that all the transactions are not typed themselves and their type is recognised by having empty or non-empty certain fields. This is the reason why a different field is used for a message call and a contract creation (i.e \textit{init} vs \textit{data}).

Let us again refer to a UML diagram in Fig. \ref{img:transaction-create-call} showing a human actor who, in the bottom part of the diagram, submits three transactions with a message call. These transactions pick-up a \textit{codeHash} from respective account and execute the smart contract (actually one of methods of the contract). This type of transaction must contain an account address in the \textit{to} field, which determines particular account contract to use when looking up the \textit{codeHash}. 

The transaction that calls a contract combines both the method signature and the input data into the \textit{data} field. The method signature is stored in a hashed (using keccak) form\cite{Howard2017methodcall}, followed by the input data. The input data are encoded in a standardized binary format -- an Abstract Binary Interface\cite{solidity2020ABI}.  This format allows for encoding both the method signature and input data in one field as a binary sequence. 

Interestingly, the ABI format is defined in the Solidity specification, which was originally intended as only one of the many languages for smart contract. However, to satisfy interoperability, alternative languages must follow the Solidity specification to communicate with other smart contracts.  We see this as a weakness and insufficient standardization, owing that Solidity neither attempts at backward compatibility, nor exposes its version as part of API. Solidity uses Semantic Versioning \cite[Chapter: Selecting a Version of Solidity]{antonopoulos2018mastering} to express a version of a deployed contract, but does not contain a means to express a required version of imported (called) contract as it has been possible for long time in systems such as OSGi\cite{alliance2003osgi}. This is nothing new that incompatibilities happen all the time \cite{raemaekers2014semantic}, but compatibility assurance does not seem to be part of Smart Contract design.

Contracts may call methods of other contracts, which is done via internal transactions. They use the same mechanism as it is used for the method call from the transaction, but internal transactions are not recorded in the blockchain.  They also do not have \textit{gasLimit} and the consumed gas is billed against the source caller, transitively when nested contracts are called. This allows for creating libraries, which are not priced themselves, but their cost must be considered when their are aggregated into other contracts.

\paragraph{v, r, s}

These three values digitally signs the transaction. Values \textit{r} and \textit{s} are computed using the ECDSA signature algorithm \cite{Fu2007ECDSA}. In the case of Ethereum an ECDSA implementation called \textit{secp256k1} \cite{Hess2000SEC2R} is in particular used. ECDSA is a symmetric function based on elliptic curves, which generates the signature out of private and public key pairs. Easily readable introduction into this function is available at blogs \cite{Knutson2018elliptic, riady2018signature}.

The private key is a secret maintained by every blockchain participant in his or her wallet \cite{ethereum2020wallet}. The wallet is a remedy persisting this key, frequently represented by a software installed on user's computers, however, more verbatim variants exist such as hardware USB keys, or even more tamper proof so called cold-storage e.g. private keys engraved on a piece of metal. 

The third value of -- \textit{v} -- is the recovery identifier, a one-byte value, which allows for obtaining a public key from the signature. When this public key is hashed by \textit{keccak}, first 20 bytes of this hash then equal to an account address.  

As it may be seen it is crucial for a user to maintain his or her private key as it servers for singing the transactions and identifies the user as the owner of an account, i.e. the crypto-currency funds. 

\subsection{Receipts Trie}

When a transaction is submitted to a blockchain, it is not executed immediately. First executes the transaction the miner, who potentially put it in the block. Then, the transaction becomes part of the blockchain and all participating nodes will sooner or later synchronize, i.e. execute the transaction. Because of this process, a user submitting originally the transaction  cannot have its results immediately. 

As a solution, result of each transaction execution is captured in the Transaction Receipt Trie.  One entry in this trie is created for each transaction and it contains at least a result code showing if the transaction has executed correctly or not. Optionally, the trie may contain log messages generated by smart contracts for more detailed debugging of contract execution. By analyzing the Receipts Trie, the user that submitted a transaction can observe when the transaction is executed and potentially study log messages to get deeper insight. 

Ethereum nodes provide a convenient API that allows the user to register for events and get notifications when a transaction receipt has arrived. Putting all this together, a user typically submit a transaction first. Since he or she cannot know when the transaction is actually executed, he or she waits for a receipt using a convenient API.

The Receipt Trie has following structure. The key is a transaction $index$ and the value is a receipt $R$. They compose a key-value pair:

%
%

\begin{equation}
RLP(index) \rightarrow RLP(R)
\end{equation}

The receipt $R$ has following fields:

\begin{itemize}
    \item Cumulative gas used in a respective block after execution of current transaction, i.e. the gas consumed by all previous transactions and the recent one. 
    \item A set of logs printed out by this transaction.
    \item A Bloom filter containing hashes of log entries to speed-up their searching. 
    \item A status code of the transaction result, expressed as a number. It semantics is application dependent. 
\end{itemize}

\paragraph{Bloom Filter}
Bloom filters had been introduced by Burton Bloom \cite{Bloom1970hash} and they speed up searching in big datasets by primary keys. The filter is a vector containing hashes of primary keys. Since the vector is usually small it may be stored e.g. in a main memory, cached, or stored close to a client in terms of distributed systems. Before looking-up the expensive storage, clients may query the filter first. Since the filter contains hashes, it may falsely claim a missing key as being present in the main data source due to hash collisions. On the other hand the filter reliably response if the key is missing and thus the hash cannot be found in the filter. It means that the filter reduces unnecessary queries in the main storage for keys that are not present. 

\paragraph{Message Logs}

As it has been mentioned, the Receipt Trie can contain a set of log messages, produced by smart contract. In the case of Solidity, the log message is emitted via a keyword \verb!emit!.  In the byte-code level, the messages are represented by EVM instructions $LOG0, \cdots, LOG4$.

The set of logs contains tuples with following items:

\begin{itemize}
    \item A contract account address, who triggered the log, i.e. executed this Smart Contract. 
    \item A sequence of 32 bytes long containing event topics.
    \item A sequence of bytes containing the log message itself. 
\end{itemize}

The log messages may contain any arbitrary data and every message ma be assigned to a topic. This way clients may listen only to messages from certain topics if they are not interested in all log messages. 

\begin{figure}[ht]
\includegraphics[width=15cm]{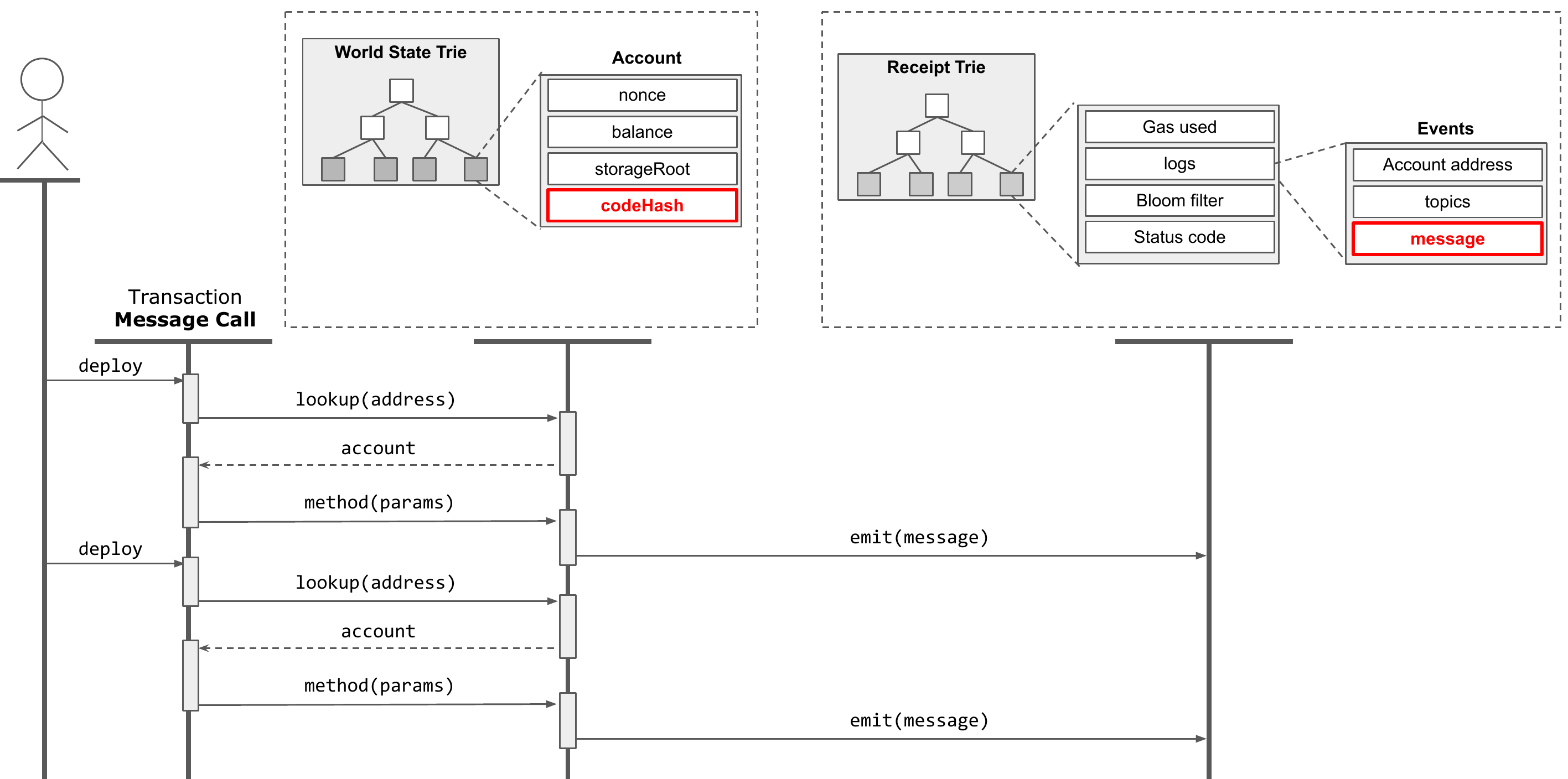}
\caption{Log Event Updates Transaction Receipt}
\label{img:transaction-receipt}
\end{figure}

We use the UML sequence diagram in Fig. \ref{img:transaction-receipt} to visualise creation of log entries. At the beginning a user must submit a transaction that executes a Smart contract, or in other words, it executes a method of Smart contract via a message call. The transaction calling a contract must contain a target account in the \textit{to} field. This field determines the account in the World State Trie, which contains the Smart Contract. Then, the contract is executed. When the contract's method contains the \verb!emit! keyword, the message propagates into the log entries.  An example snippet of Smart contract that can produce a log message may be seen in Listing \ref{sol:method-emit-msg}:

\begin{lstlisting}[language=Solidity, caption=A Method that Emits Message, label={sol:method-emit-msg}]
contract ExampleContract {
 uint storedData
 event Sent(msg);
 constructor() public {}

 function method(uint x) public { 
   storeData = x
   emit Sent('Success'); 
  } 
}
\end{lstlisting}

\subsection{World State Trie}
\label{sec:world-state-trie}

In contrast to Transaction Tries, the World State Trie is a mutable data-structure capturing the most recent state of the blockchain. World State Trie contains a mapping between addresses and accounts, i.e. paths in this trie link records with account information. An address of an account is a 160-bit (20 bytes) identifiers, which is created as first 20 bytes of a public key from the user signature. 

Ethereum knows two types of Accounts: Externally Owned and Contract Account. Their form is the same, both are leaves in the World State Trie, but they can be distinguished by empty or non-empty values in certain fields. 

As it was mentioned, every account $A$ stored is mapped via an account address. The address is furthermore hashed (detailed in Section \ref{sec:secure-trie}), which creates a key-value pair:

\begin{equation}
    keccak(address) \rightarrow RLP(A)
\end{equation}

The structure $A$ contains following fields: 
\begin{itemize}
    \item \textbf{nonce}: A scalar, for Externally Owned Accounts it is equivalent to the number of transactions sent from this address; for Contract Accounts it contains number of creation of the contract.
    \item \textbf{balance}: A scalar, contains funds available in this account.
    \item \textbf{storageRoot}: A 256-bit hash, containing a hash of root of the Account Storage Trie. Empty for Externally owned accounts. 
    \item \textbf{codeHash}: Hash of the EVM code. Particular EVM byte-code is stored in an underlying database under this hash as  a key. Empty for Externally owned accounts. 
\end{itemize}

The Externally owned account is meant for a user and it holds a balance of his funds. Although these accounts are often managed by software, their purpose is to serve human end users to store their funds, similarly to standard bank accounts.  External accounts have the fields \textit{storageRoot} and \textit{codeHash} empty.

The Contract Account contains a Smart Contract and it links contract byte-code via the filed  \textit{codeHash}. While this field contains only a hash (i.e. keccak), the contract's byte-code must be looked up from the database this hash being a key. Furthermore, this account contains \textit{storageRoot}, which is an additional trie structure:  Account Storage Trie persisting contract data. It will be described independent in the next section. 

The World State Trie is manipulated via transitions. Obviously, the transaction type must match with the account type. It is, for instance, not possible to send a transaction encoding a message call but actually pointing to an Externally owned account. 

When a transaction addresses an Externally Owned Account, the balance of this account is updated as prescribed in the transaction. As endless flow of transactions can modify the same account, the World State Trie is an always changing structure. 

When a transaction address a Smart Contract account, the contract stored in \textit{codeHash} is executed according to \textit{data} filed of the contract. Part of the contract execution is typically a change of state, or in other words updates of contract's global variables. It triggers updates in the State Trie referred via \textit{storageRoot}. For this reason the State Trie is an always modified structure as well.

\subsection{UTXO Based and Account Based Blockchains}

The concept of managing accounts in Ethereum is different from another blockchain technology, namely Bitcoin\cite{nakamoto2019bitcoin}.  Bitcoin does not accumulate balances of users in accounts, but instead it recognises only transactions, which contain addresses and crypto-currency amounts. Transactions transfer balances between these addresses and the balance must be transferred as a whole. It is called unspent transaction output -- UTXO. If a user wants to pay for something which is priced more than the amount the user has on one address, he or she must combine amounts from more addresses (assuming the user owns more addresses). If the sum does not exactly match with the price to pay, the user must combine the amounts to sent a higher value. Furthermore, the user provides a so called change address, which receives a change after the price is paid.

UTXO more remains of real-life payments where people exchange whole coins or notes and get change back. This system has its pros and cons as discussed informally on blogs \cite{ConsenSys2016UTXO}. Interoperatibility of these two systems were analysed by researchers as well  \cite{zahnentferner2018chimeric}.
One of the benefit of UTXO is that the transactions can run in parallel as there is no dependency between them. UTXO can also scale better for similar reasons, because no data must be looked-up to update global balances of accounts. The addresses are stored in user wallets and users only create a transaction out of these addresses, which get delivered to other parties. No previous balances must be located to update them.  On the other hand it is impossible to easily get user balances e.g. in Bitcoin wallets. The Wallets must instead track all the incoming addresses and their balances to produce a summary. In other words, while Ethereum transactions may deduct certain amount from the account directly, Bitcoin and other UTXO based systems may transfer only the whole amount and receive exchange.  

\subsection{Storage Trie}

The World State Trie refers to Smart Contract data via an account and its field \textit{storageRoot}, which is a hash of the root of the Storage Trie. This trie is directly modified via a Smart Contract byte-code using instructions SLOAD and SSTORE. 

Every key in this trie is an $index$  of a $slot$ stored in the leaf node. The key is furthermore  hashed. The index represent one or more global variables in the Smart Contract. It is up-to the compiler to determine an index for every variable at compile time. Furthermore, the index is hashed. We can formulate that index and slot pairs are stored in the trie as: 

\begin{equation}
    keccak(index) \rightarrow RLP(slot)
\end{equation}

\begin{figure}[ht]
\includegraphics[width=12cm]{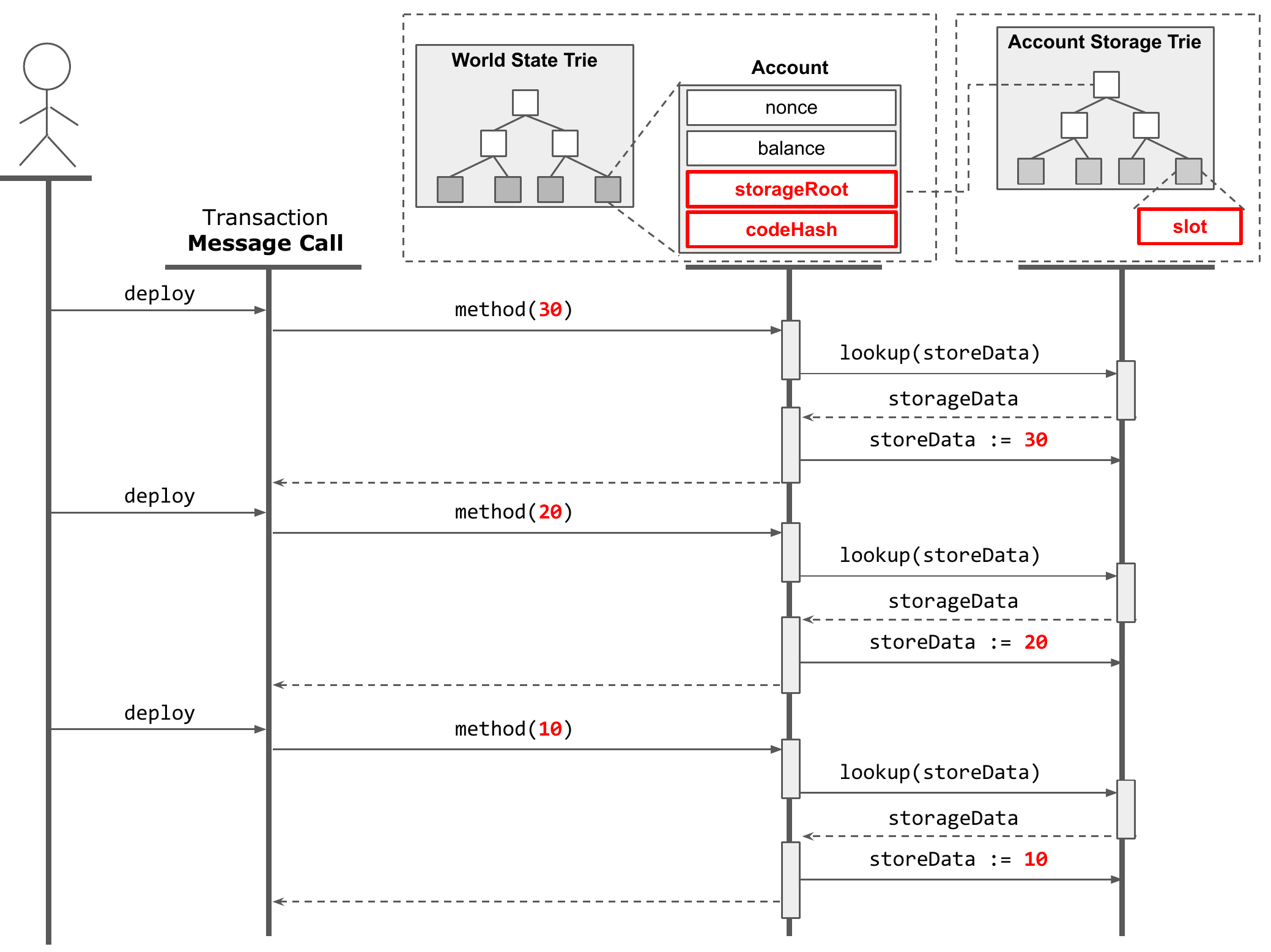}
\caption{Storage Trie Updates}
\label{img:transaction-storage-trie}
\end{figure}

Let us illustrate how this trie is modified with the already used Listing \ref{sol:method-emit-msg} and an UML diagram in Fig. \ref{img:transaction-storage-trie}. Let us assume that a user has already submitted transactions executing this Smart Contract. Let us also assume that the transaction field \textit{data} has contained signature of the function called \verb!method!, and this function has been called three times with values $30$, $20$, and $10$. This process is depicted in the UML sequence diagram, where an account addressed in the transaction is located first, then the contract from the field \textit{codeHash} is executed and the Account Storage Trie linked via \textit{storageRoot} is updated.  Since the Smart in  Listing \ref{sol:method-emit-msg} contains one global variable \verb!storedData! and this variable is modified via the executed method, the Account Storage Trie is sequentially modified as the transactions are processed. 

One leaf in the Storage Trie is a slot, which contain one or more variables. The slot is wide 32 bytes and its content depends on a data type of particular variable, furthermore detailed in Solidity compiler documentation\cite{solidity2020storage}. 

It is again interesting to see that the data format valid throughout the blockchain is defined in Solidity, which is however only one possible implementation. In other words, if an alternative compiler stores the data differently, it will produce incompatible data storage. The storage itself does not contain any versioning information, which means that even Solidity cannot evolve to store the data differently. The only barrier of incompatible changes seems to be at the moment that the contract code cannot be updated. 

We provide here an overview of variable format produced by current Solidity:

\paragraph{Statically Sized Variables}
For statically-sized variables, multiple items are grouped in one slot up to total of 32 bytes. The first value which does not fit into 32 bytes starts a new slot.  Structures and arrays start a new slot and occupy whole slots, except that their items are packed as well. First variable spotted by the compiler is assigned index $\mathrm{0x0}$, every new slot uses the next available indexes. 

\begin{figure}[ht]
\includegraphics[width=8cm]{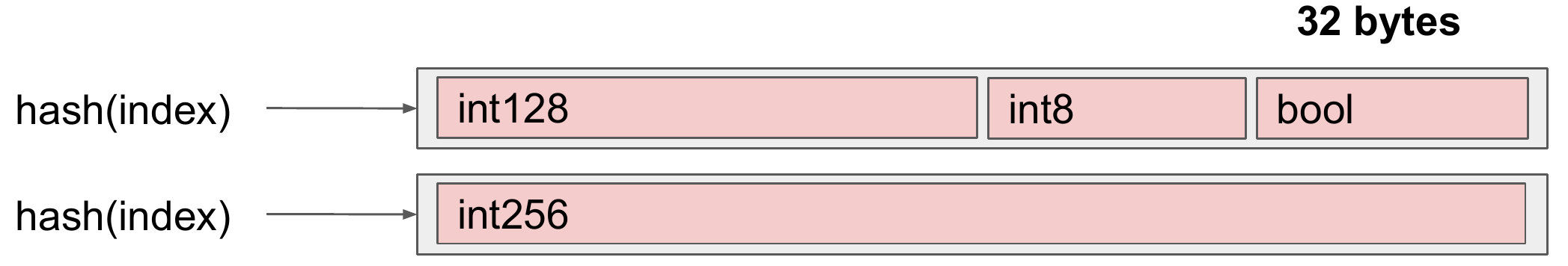}
\caption{Statically Sized Variables}
\label{img:static-variables}
\end{figure}

An example in Fig. \ref{img:static-variables} shows one slot with three packed variables (\verb!int128!, \verb!int8! and one \verb!bool!), and one wide (\verb!int256!) variable that cannot be packed.

When a contract uses inheritance, its slots are shifted in the storage to a next available slot after fitting-in state variables from parents.

\begin{figure}[ht]
\includegraphics[width=8cm]{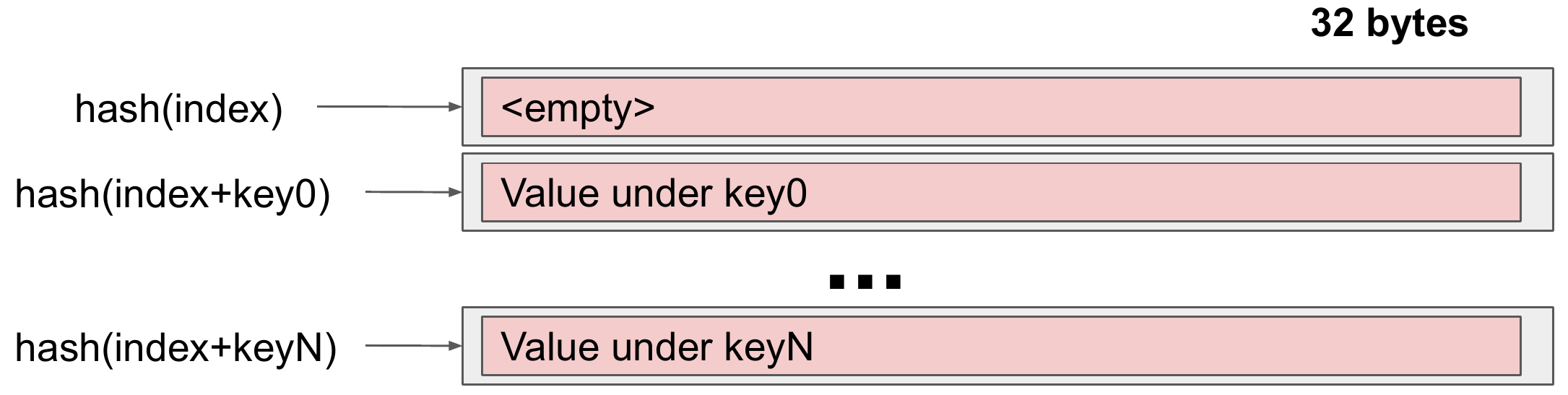}
\caption{Map Variables}
\label{img:map-variables}
\end{figure}

\paragraph{Maps}
In the case of maps, a first available slot's $index$ is selected. For every key in the map, the key is concatenated with this slot index and the hash is computed: 
\begin{equation}
    index_{val} = keccak(index + key)
\end{equation}
Note that $+$ here means string concatenation. This index points to a slot with the map value. Notice that the slot under the first $index$ remains itself empty. An example with a map starting at \verb!index! is provided in Fig. \ref{img:map-variables}.

\begin{figure}[ht]
\includegraphics[width=8cm]{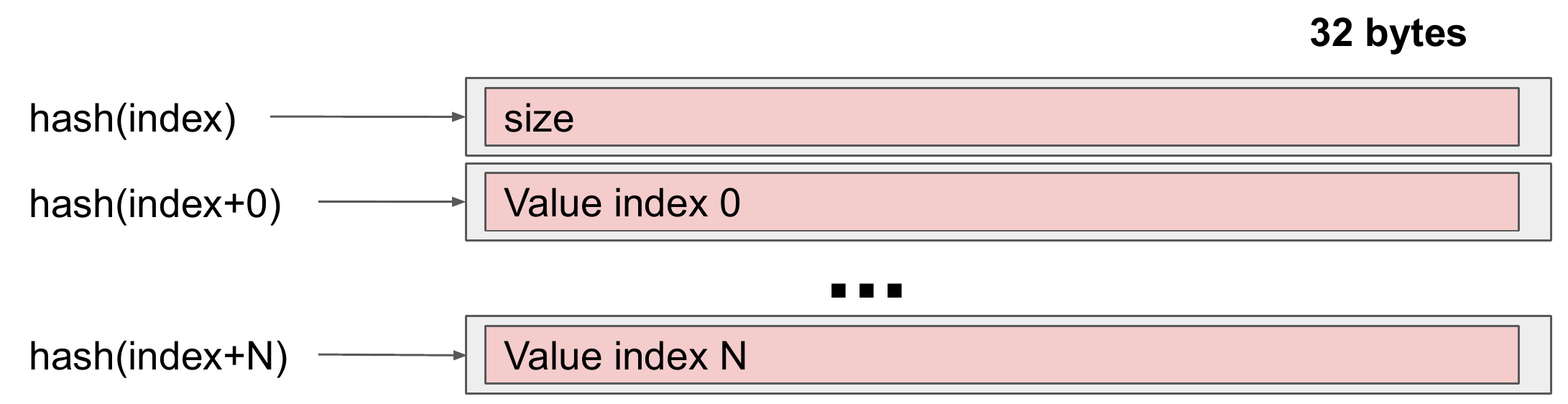}
\caption{Dynamic Arrays}
\label{img:array-variables}
\end{figure}

\paragraph{Dynamic Arrays} The next available $index$ is associated with the array. This index points to a slot which contains the array size (it is not empty as for the maps). The hash of $index$ points to the slot with the beginning of the array and all array items are laid out sequentially from this index.  To get a value from the array, it must be computed:
\begin{equation}
    index_i = keccak(index + i)    
\end{equation}
where $index_i$ points to a slot with the value from array index $i$.  As it may be seen, arrays and maps are stored fundamentally the same. However, the arrays have additional benefit that they also pack values if they fit more variables into one slot. An example  with an array starting at \verb!index! is shown in Fig. \ref{img:array-variables}.

\begin{figure}[ht]
\includegraphics[width=8cm]{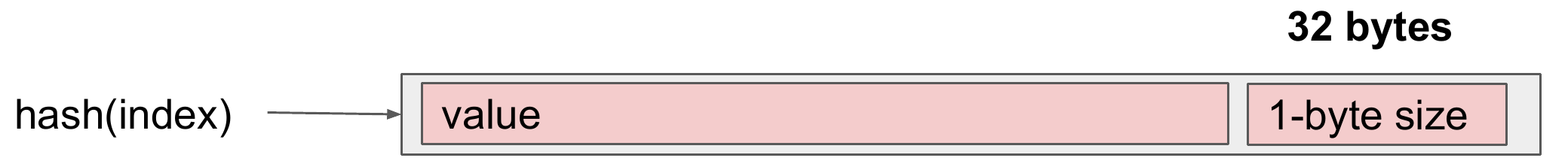}
\caption{Byte Arrays and String}
\label{img:string-variables}
\end{figure}

\paragraph{Byte Arrays and String}
When these types are shorter then 32 bytes, they are stored in one slot so that one byte contains the size and 31 bytes are used for data. For longer strings or arrays, they are stored the same way as dynamic arrays. One slot contains the length and following slots the payload.  An example of a short string is shown in Fig. \ref{img:string-variables}.

\subsection{Authenticated Storage}

Merkle Patricia Tries are used across the blockchain among other reasons to provide a so called authenticated storage. This is done by employing Merkle hashes. Once any data subset is obtained from the blockchain, it always come with its hash. This hash essentially authenticates, i.e. proves validity of the data. 

\begin{figure}[ht]
\includegraphics[width=8cm]{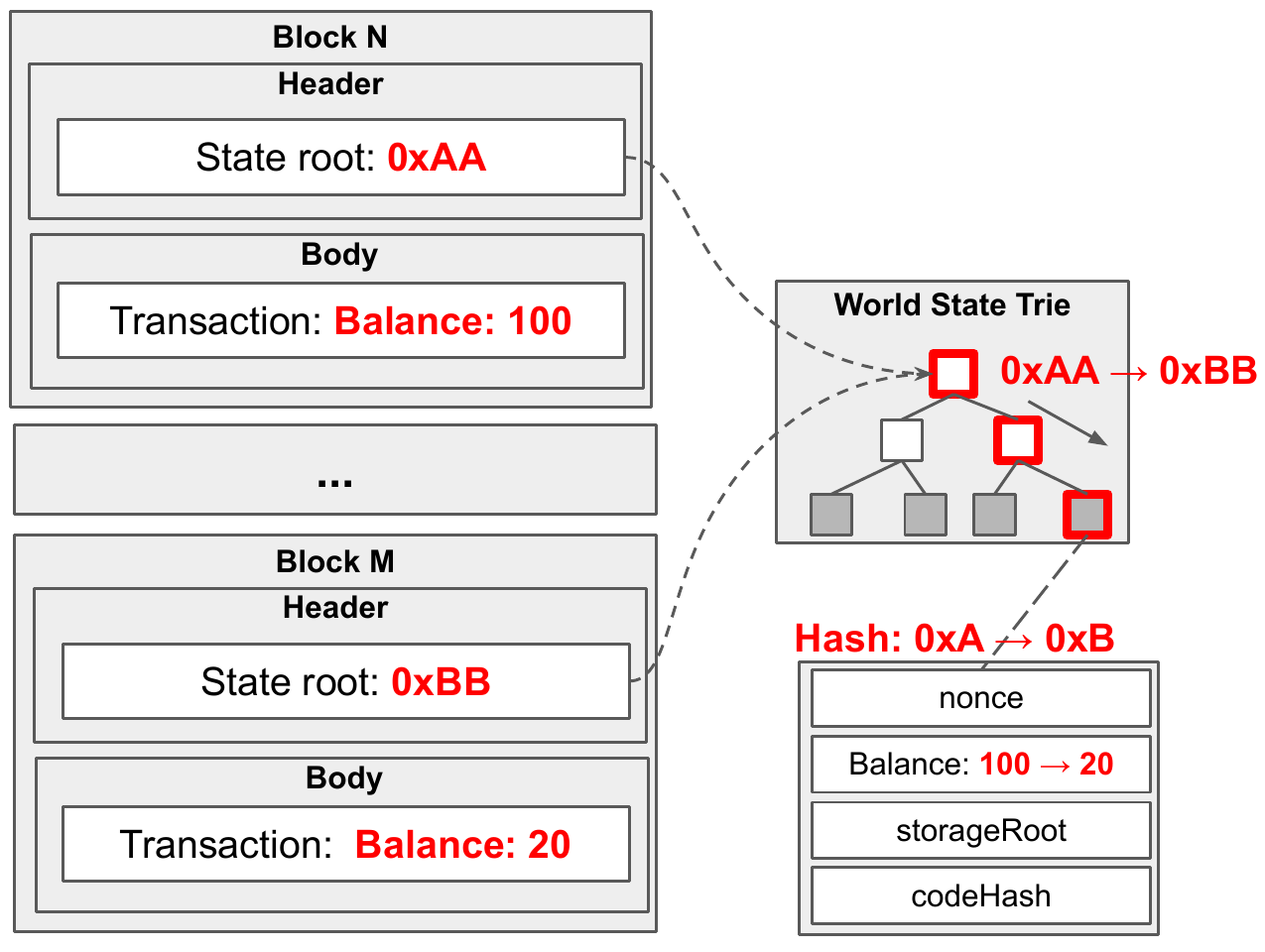}
\caption{Authenticated Storage}
\label{img:authenticate-storage}
\end{figure}

This concept may be furthermore illustrated on an example in Fig. \ref{img:authenticate-storage}. Let us assume there is a block with a transaction that sets an account balance to `100', visualised as \textit{Block N}. This transaction updates an account in the World State Trie first. Then, the hash of this account (an example value $\mathrm{0xA}$ is shown) is computed and all the hashes up to the root are recomputed. It creates a hash (example value $\mathrm{0xAA}$), which is stored in the header of the \textit{Block N}. Any time later, another block contains a transaction that changes a balance to a value `20'. This changes the hash of respective account and recomputes all hashes up to the root (example hashes $\mathrm{0xB}$ and $\mathrm{0xBB}$ are used). This new hash is stored in the header of the block \textit{Block M}, and the hash in the block \textit{Block N} essentially becomes obsolete.

The result is that both the data structure and the blockhain are bound to these hashes. If someone, for instance, manipulates the datasource in an attempt to prove he is holding different funds, it may be easily verified that the hash of this corrupted (or obsolete) datasource does not match with the latest block header. 

If someone, in contrary, tries to use for instance an older block to prove his holding of certain funds, it may be again easily verified that the block header does not match with the most recent dataset.

\subsection{Pseudo Structure: Secure Trie}
\label{sec:secure-trie}

Secure Trie is a wrapper of another trie. This is actually not an independent data structure, but it is rather a term to say that keys (paths) in the Trie are hashed as well. We mentioned this term here as it is used in Geth and Parity implementation and also on a couple of blogs\cite{Ethereum2020rationale,Kim2020securetrie}.
Practically, Ethereum employs a Secure Trie for World State Trie and Account Storage Trie. 

For instance for an account address, the key is actually not a 20 bytes address, but 32 bytes hash, when it is stored in the trie. It means that whenever there is an account address, this address must be hashed first, then it may be looked-up or stored in the trie.  Similarly, a variable identifier for a value from Accout Storage Trie is hashed. It means that every identifier populated to SLOAD and SSTORE instructions is hashed first, then it is accessed in the trie. 

The rationale behind this is to prevent DOS attacks. An attacker could construct a payload (e.g. account addresses) in a way that degrades a trie to contain long paths with minimal branching. It would lead to an explosion of look-ups in respective tries. When the key is hashed, it randomises its string representation and contributes to more random distribution of branching. 

\section{CONCLUSIONS}

\bibliographystyle{plain} 
\bibliography{literature}

\begin{thebibliography}{10}

\bibitem{afanasev2017performance}
Maxim~Ya Afanasev, Yuri~V Fedosov, Anastasiya~A Krylova, and Sergey~A
  Shorokhov.
\newblock Performance evaluation of the message queue protocols to transfer
  binary json in a distributed cnc system.
\newblock In {\em 2017 IEEE 15th International Conference on Industrial
  Informatics (INDIN)}, pages 357--362. IEEE, 2017.

\bibitem{alliance2003osgi}
OSGi Alliance.
\newblock {\em Osgi service platform, release 3}.
\newblock IOS press, 2003.

\bibitem{Anjana2019efficient}
P.~S. {Anjana}, S.~{Kumari}, S.~{Peri}, S.~{Rathor}, and A.~{Somani}.
\newblock An efficient framework for optimistic concurrent execution of smart
  contracts.
\newblock In {\em 2019 27th Euromicro International Conference on Parallel,
  Distributed and Network-Based Processing (PDP)}, pages 83--92, Feb 2019.

\bibitem{antonopoulos2018mastering}
Andreas~M Antonopoulos and Gavin Wood.
\newblock {\em Mastering ethereum: building smart contracts and dapps}.
\newblock O'reilly Media, 2018.

\bibitem{bashir2017mastering}
Imran Bashir.
\newblock {\em Mastering blockchain}.
\newblock Packt Publishing Ltd, 2017.

\bibitem{Bloom1970hash}
Burton~H. Bloom.
\newblock Space/time trade-offs in hash coding with allowable errors.
\newblock {\em Commun. ACM}, 13(7):422–426, July 1970.

\bibitem{borthakur2013under}
Dhruba Borthakur.
\newblock Under the hood: Building and open-sourcing rocksdb.
\newblock {\em Facebook Engineering Notes}, 2013.
\newblock [Online; accessed 3-March-2020].

\bibitem{ConsenSys2016UTXO}
ConsenSys.
\newblock Utxo.
\newblock
  url={https://medium.com/@ConsenSys/thoughts-on-utxo-by-vitalik-buterin-2bb782c67e53}.
\newblock [Online; accessed 3-March-2020].

\bibitem{Dannen2017introducing}
Chris Dannen.
\newblock {\em Introducing Ethereum and Solidity}, volume~1.
\newblock Springer, 2017.

\bibitem{Briandais1959file}
Rene De~La~Briandais.
\newblock File searching using variable length keys.
\newblock In {\em Papers Presented at the the March 3-5, 1959, Western Joint
  Computer Conference}, IRE-AIEE-ACM ’59 (Western), page 295–298, New York,
  NY, USA, 1959. Association for Computing Machinery.

\bibitem{Dickerson2017concurrency}
Thomas Dickerson, Paul Gazzillo, Maurice Herlihy, and Eric Koskinen.
\newblock Adding concurrency to smart contracts.
\newblock In {\em Proceedings of the ACM Symposium on Principles of Distributed
  Computing}, page 303–312, New York, NY, USA, 2017. Association for
  Computing Machinery.

\bibitem{Ethereum2020rationale}
Ethereum.
\newblock Ethereum: Design rationale.
\newblock url={https://github.com/ethereum/wiki/wiki/design-rationale}.
\newblock [Online; accessed 3-March-2020].

\bibitem{ethereum2020wallet}
Ethereum.
\newblock Ethereum wallets.
\newblock url={https://ethereum.org/wallets/}, journal={ethereum.org}.
\newblock [Online; accessed 3-March-2020].

\bibitem{ethereum2020ethahs}
Ethereum.
\newblock ethereum/wiki.
\newblock url={https://github.com/ethereum/wiki/wiki/Ethash}.
\newblock [Online; accessed 3-March-2020].

\bibitem{geth}
Go~Ethereum.
\newblock Geth.
\newblock url={https://geth.ethereum.org}.
\newblock [Online; accessed 3-March-2020].

\bibitem{Fredkin1960trie}
Edward Fredkin.
\newblock Trie memory.
\newblock {\em Commun. ACM}, 3(9):490–499, September 1960.

\bibitem{Fu2007ECDSA}
David~E. Fu and Jerome~A. Solinas.
\newblock Ike and ikev2 authentication using the elliptic curve digital
  signature algorithm (ecdsa).
\newblock {\em RFC}, 4754:1--15, 2007.

\bibitem{etherchain202hardforks}
Bitfly gmbh.
\newblock Ethereum hard forks.
\newblock url={https://www.etherchain.org/hardForks}.
\newblock [Online; accessed 3-March-2020].

\bibitem{Hefele2019conceptual}
Alexander Hefele.
\newblock A conceptual model for ethereum blockchain analytics.
\newblock Master's thesis, Technishe Universitat Munchen, 2 2019.

\bibitem{Hess2000SEC2R}
Philipp Hess.
\newblock Sec 2: Recommended elliptic curve domain parameters.
\newblock In {\em STANDARDS FOR EFFICIENT CRYPTOGRAPHY}, 2000.

\bibitem{Howard2017methodcall}
Howard.
\newblock How to decipher a smart contract method call.
\newblock
  url={https://medium.com/@hayeah/how-to-decipher-a-smart-contract-method-call-8ee980311603}.
\newblock [Online; accessed 3-March-2020].

\bibitem{Howard2017contractcreation}
Howard.
\newblock The smart contract creation process.
\newblock
  url={https://medium.com/@hayeah/diving-into-the-ethereum-vm-part-5-the-smart-contract-creation-process-cb7b6133b855}.
\newblock [Online; accessed 3-March-2020].

\bibitem{Humiston2018thedao}
Pete Humiston.
\newblock Smart contract attacks [part 1] - 3 attacks we should all learn from
  the dao.
\newblock
  url={https://hackernoon.com/smart-contract-attacks-part-1-3-attacks-we-should-all-learn-from-the-dao-909ae4483f0a}.
\newblock [Online; accessed 3-March-2020].

\bibitem{Kasireddy2017howethworks}
Preethi Kasireddy.
\newblock How does ethereum work, anyway?
\newblock
  url={https://medium.com/@preethikasireddy/how-does-ethereum-work-anyway-22d1df506369
  }.
\newblock [Online; accessed 3-March-2020].

\bibitem{kim2019ethanos}
Jae-Yun Kim, Jun-Mo Lee, Yeon-Jae Koo, Sang-Hyeon Park, and Soo-Mook Moon.
\newblock Ethanos: Lightweight bootstrapping for ethereum.
\newblock {\em arXiv preprint arXiv:1911.05953}, 2019.

\bibitem{Kim2020securetrie}
Seung~Woo Kim.
\newblock Secure tree — why state trie’s key is 256 bits.
\newblock
  url={https://medium.com/codechain/secure-tree-why-state-tries-key-is-256-bits-1276beb68485}.
\newblock [Online; accessed 3-March-2020].

\bibitem{Knuth1998art3}
Donald~E. Knuth.
\newblock {\em The Art of Computer Programming, Volume 3: (2nd Ed.) Sorting and
  Searching}.
\newblock Addison Wesley Longman Publishing Co., Inc., USA, 1998.

\bibitem{Knutson2018elliptic}
Hans Knutson.
\newblock What is the math behind elliptic curve cryptography?
\newblock
  url={https://hackernoon.com/what-is-the-math-behind-elliptic-curve-cryptography-f61b25253da3}.
\newblock [Online; accessed 3-March-2020].

\bibitem{kolb2020core}
John Kolb, Moustafa AbdelBaky, Randy~H. Katz, and David~E. Culler.
\newblock Core concepts, challenges, and future directions in blockchain: A
  centralized tutorial.
\newblock {\em ACM Comput. Surv.}, 53(1), February 2020.

\bibitem{lao2020survey}
Laphou Lao, Zecheng Li, Songlin Hou, Bin Xiao, Songtao Guo, and Yuanyuan Yang.
\newblock A survey of iot applications in blockchain systems: Architecture,
  consensus, and traffic modeling.
\newblock {\em ACM Comput. Surv.}, 53(1), February 2020.

\bibitem{leveldb2014fast}
A~LevelDB.
\newblock Fast and lightweight key/value database library by google.
\newblock url={https://github.com/google/leveldb}, 2014.
\newblock [Online; accessed 3-March-2020].

\bibitem{Merkle1987tree}
Ralph~C. Merkle.
\newblock A digital signature based on a conventional encryption function.
\newblock In Carl Pomerance, editor, {\em Advances in Cryptology --- CRYPTO
  '87}, pages 369--378, Berlin, Heidelberg, 1988. Springer Berlin Heidelberg.

\bibitem{Morrison1968patricia}
Donald~R. Morrison.
\newblock Patricia—practical algorithm to retrieve information coded in
  alphanumeric.
\newblock {\em J. ACM}, 15(4):514–534, October 1968.

\bibitem{nakamoto2019bitcoin}
Satoshi Nakamoto.
\newblock Bitcoin: A peer-to-peer electronic cash system.
\newblock Technical report, Manubot, 2019.

\bibitem{Olszewicz2019hard}
Josh Olszewicz.
\newblock Ethereum price analysis - istanbul hard fork scheduled for next week.
\newblock
  url={https://bravenewcoin.com/insights/ethereum-price-analysis-istanbul-hard-fork-scheduled-for-next-week}.
\newblock [Online; accessed 3-March-2020].

\bibitem{oracle2020RMI}
Oracle.
\newblock Designing a remote interface.
\newblock url={https://docs.oracle.com/javase/tutorial/rmi/designing.html}.
\newblock [Online; accessed 3-March-2020].

\bibitem{patsonakis2019alternative}
Christos Patsonakis and Mema Roussopoulos.
\newblock An alternative paradigm for developing and pricing storage on smart
  contract platforms.
\newblock In {\em 2019 IEEE International Conference on Decentralized
  Applications and Infrastructures (DAPPCON)}, pages 170--175. IEEE, 2019.

\bibitem{Bertoni2011Keccak}
G.~Bertoni{,} J. Daemen{,}~M. Peeters and G.~Van Assche.
\newblock {The Keccak reference}.
\newblock Round 3 submission to NIST SHA-3, 2011.

\bibitem{raemaekers2014semantic}
Steven Raemaekers, Arie Van~Deursen, and Joost Visser.
\newblock Semantic versioning versus breaking changes: A study of the maven
  repository.
\newblock In {\em 2014 IEEE 14th International Working Conference on Source
  Code Analysis and Manipulation}, pages 215--224. IEEE, 2014.

\bibitem{raju2018mlsm}
Pandian Raju, Soujanya Ponnapalli, Evan Kaminsky, Gilad Oved, Zachary Keener,
  Vijay Chidambaram, and Ittai Abraham.
\newblock mlsm: Making authenticated storage faster in ethereum.
\newblock In {\em 10th $\{$USENIX$\}$ Workshop on Hot Topics in Storage and
  File Systems (HotStorage 18)}, 2018.

\bibitem{riady2018signature}
Yos Riady.
\newblock Signing and verifying ethereum signatures.
\newblock url={https://yos.io/2018/11/16/ethereum-signatures/}, Nov 2018.
\newblock [Online; accessed 3-March-2020].

\bibitem{sanguna2018proposals}
Carlos Sangu{\~n}a, Sebasti{\'a}n Suarez, Antonio Russoniello, Miguel~A Astor,
  and Wilmer Pereira.
\newblock Proposals for a smart contracts platform for cryptocurrencies based
  on the mini-blockchain scheme.

\bibitem{solidity2020ABI}
Solidity.
\newblock Contract {ABI} specification.
\newblock url={https://solidity.readthedocs.io/en/v0.6.2/abi-spec.html}.
\newblock [Online; accessed 3-March-2020].

\bibitem{solidity2020language}
Solidity.
\newblock Solidity language.
\newblock url={https://solidity.readthedocs.io/en/v0.6.3/}.
\newblock [Online; accessed 3-March-2020].

\bibitem{solidity2020storage}
Solidity.
\newblock Storage layout.
\newblock
  url={https://solidity.readthedocs.io/en/develop/miscellaneous.html\#layout-of-state-variables-in-storage}.
\newblock [Online; accessed 3-March-2020].

\bibitem{Sompolinsky2015secure}
Yonatan Sompolinsky and Aviv Zohar.
\newblock Secure high-rate transaction processing in bitcoin.
\newblock In Rainer B{\"o}hme and Tatsuaki Okamoto, editors, {\em Financial
  Cryptography and Data Security}, pages 507--527, Berlin, Heidelberg, 2015.
  Springer Berlin Heidelberg.

\bibitem{parity}
Parity Technologies.
\newblock Parity.
\newblock url={https://www.parity.io}.
\newblock [Online; accessed 3-March-2020].

\bibitem{Voulgari2019ethereum}
Athina Voulgari.
\newblock Etherum analytics.
\newblock Master's thesis, ETH Zurich, 6 2019.

\bibitem{wood2019ethereum}
Gavin Wood.
\newblock Ethereum: A secure decentralised generalised transaction ledger
  byzantium version 7e819ec - 2019-10-20, 2019.
\newblock Accessed: 2020-02-18.

\bibitem{Wst2019ACEAA}
Karl W{\"u}st, Sinisa Matetic, Silvan Egli, Kari Kostiainen, and Srdjan Capkun.
\newblock Ace: Asynchronous and concurrent execution of complex smart
  contracts.
\newblock {\em IACR Cryptology ePrint Archive}, 2019:835, 2019.

\bibitem{zahnentferner2018chimeric}
Joachim Zahnentferner.
\newblock Chimeric ledgers: Translating and unifying utxo-based and
  account-based cryptocurrencies.
\newblock {\em IACR Cryptology ePrint Archive}, 2018:262, 2018.

\bibitem{zhang2018enabling}
An~Zhang and Kunlong Zhang.
\newblock Enabling concurrency on smart contracts using multiversion ordering.
\newblock In {\em Asia-Pacific Web (APWeb) and Web-Age Information Management
  (WAIM) Joint International Conference on Web and Big Data}, pages 425--439.
  Springer, 2018.

\bibitem{zhang2019security}
Rui Zhang, Rui Xue, and Ling Liu.
\newblock Security and privacy on blockchain.
\newblock {\em ACM Comput. Surv.}, 52(3), July 2019.

\bibitem{zhu2020applications}
Qingyi Zhu, Seng~W. Loke, Rolando Trujillo-Rasua, Frank Jiang, and Yong Xiang.
\newblock Applications of distributed ledger technologies to the internet of
  things: A survey.
\newblock {\em ACM Comput. Surv.}, 52(6), November 2019.

\end{thebibliography}

\end{document}